\documentclass[]{hdsr}
\usepackage{amsmath}
\usepackage{amssymb}
\DeclareUnicodeCharacter{00A0}{ }

\DeclareMathOperator*{\argmin}{arg\,min}
\usepackage{color}
\graphicspath{{figs/}}
\usepackage{xcolor}

\begin{document}

% Larger bottom margin for the first page
\newgeometry{bottom=1.5in}

% Editorial staff will replace the following values:
% 1. Volume number
% 2. Issue number
% 3. Article DOI
% e.g. for Volume 2, Issue 3, DOI 12.345:
% \volumeheader{2}{3}{12.345}
%\volumeheader{0}{0}{00.000}

\begin{center}

  \title{Effective Generative AI: The Human-Algorithm Centaur}
  \maketitle

  % Start page numbering on second page. Must appear *after* \maketitle
  \thispagestyle{empty}
  
  \vspace*{.2in}

  % Authors and Affiliations
  \begin{tabular}{cc}
    Soroush Saghafian\upstairs{\affilone,\affiltwo,\affilthree,\affilfour,\affilfive,\affilsix,\affilseven,\affileight,*}, Lihi Idan\upstairs{\affilnine}\vspace{10mm}
   \\[0.25ex]
   
   {\small \upstairs{\affilone} Associate Professor, Harvard University} \\ {\small \upstairs{\affiltwo} Visiting Scholar, MIT} \\ {\small \upstairs{\affilthree} Faculty Affiliate, Harvard Data Science Initiative} \\ {\small \upstairs{\affilfour} \small Founder and Director, Harvard’s Public Impact Analytics Science Lab (PIAS-Lab)} \\ {\small \upstairs{\affilfive} \small Core Faculty, Harvard Center for Health Decision Science} \\\ {\small \upstairs{\affilsix} \small Faculty Affiliate, Harvard Mossavar-Rahmani Center for Business and Government} \\ {\small \upstairs{\affilseven} \small Faculty Affiliate, Belfer Center for Science \& International Affairs} \\ {\small \upstairs{\affileight} \small Faculty Affiliate, Center for Public Leadership} \\
   {\small \upstairs{\affilnine} \small Postdoctoral Fellow, Harvard University} \\
  \end{tabular}
  
  % Replace with corresponding author email address
  \emails{
    \upstairs{*}Corresponding Author (soroush\_saghafian@hks.harvard.edu) 
    }
  \vspace*{0.4in}

\begin{abstract}
Advanced analytics science methods have enabled combining the power of artificial and human intelligence, creating \textit{centaurs} that allow superior decision-making. Centaurs are hybrid human-algorithm models that combine both formal analytics and human intuition in a symbiotic manner within their learning and reasoning process. We argue that the future of AI development and use in many domains needs to focus more on centaurs as opposed to other AI approaches. This paradigm shift towards centaur-based AI methods raises some fundamental questions: How are centaurs different from other human-in-the-loop methods? What are the most effective methods for creating centaurs? When should centaurs be used, and when should the lead be given to pure AI models? Doesn't the incorporation of human intuition---which at times can be misleading---in centaurs' decision-making process degrade its performance compared to pure AI methods?  This work aims to address these fundamental questions, focusing on recent advancements in generative AI, and especially in Large Language Models (LLMs), as a main case study to illustrate centaurs' critical essentiality to future AI endeavors. 
\end{abstract}
\end{center}

\vspace*{0.15in}
\hspace{10pt}
  
\copyrightnotice

\section{Introduction}

Developing analytics science methods that can enable combining the power of artificial and human intelligence has brought the concept of \textit{centaurs} from myth to reality. In the Greek mythology, centaurs are half-human and half-horse creatures. In modern analytics science, {centaurs represent strong merges between humans and algorithms, referring to hybrid human-algorithm models that combine both formal analytics and human intuition in a symbiotic manner.} One of the main users in the U.S. has been the Defense Department, which has been working with tech companies to combine the power of algorithms with the capabilities of humans \citep{1}. The concept has attracted the attention of the U.S. military, both in research programs at the Defense Advanced Research Projects Agency and the Pentagon’s third-offset strategy for military advantage \citep{2}. Robert O. Work, for example, who was the deputy secretary of defense under Presidents Trump and Barack Obama, advocated for the idea of centaur weapons systems, which would require human control, instead of pure Artificial Intelligence (AI) systems, and could combine the power of AI with the capabilities of humans \citep{3}.

The concept of centaurs is not that new, but it received spotlight attention within the analytics science domain because of its success in applications like playing free-style chess. Specifically, prominent advocates of free-style chess like Gary Kasparov repeatedly argued that humans paired with algorithms can do better than just the single strongest computer program in chess \citep{4}. As the chess legend put it:

\vspace{10pt}
“Weak human plus machine plus better process was superior to a strong computer alone and, more remarkably, superior to a strong human plus machine plus inferior process.” \citep{5}

\vspace{10pt}
Beyond free-style chess, the centaur model is being widely used in a variety of applications of analytics science. In clinical decision-making related to rehabilitation assessment, for example, algorithms provide therapists with detailed analysis on patient’s status, where the collaboration with therapist and such algorithm is shown to improve the practices of rehabilitation assessment \citep{6}. 

Research in the first author's lab at Harvard, which was conducted in collaboration with the Mayo Clinic, showed very promising results for a centaur model developed to enhance decision-making and reduce readmission risks for a large number of patients who underwent transplantation.  The research showed that combining human experts’ intuition with the power of a strong machine-learning algorithm through a human-algorithm centaur model can outperform both the best algorithm and the best human experts \citep{sor}.

Other examples of using the centaur model to create public impact include systems for spotting anomalies and preventing cyber-attacks, improving design components in manufacturing systems, and assisting officers balance their workloads and helping them to better ensure public safety \citep{2}. And the potential for developing and making use of centaurs is endless. Thus, it is reasonable to expect most data-driven organizations to take advantage of them in the near future. A department of human services, for example, can use algorithms to help predict which child welfare cases are likely to lead to child fatalities and raise a red flag for high-risk cases. Such cases are then reviewed by human experts and the results are shared with frontline staff, who then might choose remedies designed to lower risk and improve outcomes \citep{8}. The algorithm then can be augmented by using human intuition related to specific cases, creating a human-algorithm centaur. 

In this article, we focus on the role of centaurs in recent advancements in Generative AI, and especially in Large Language Models (LLMs). We first present a framework that allows an understanding of the core
characteristics of centaurs. We argue that the incorporation of {\em human intuition} and {\em symbiotic
learning}---a specific learning method based on a human-algorithm symbiosis that we will define
and theorize---are key characteristics of centaurs that distinguish them from other AI models. Using these core characteristics, we also present a few specific methods of creating centaurs. We
then argue that the growth and success of LLMs are to a great extent due to the fact that they
are moved from pure AI algorithms to human-algorithm centaurs. We present various pieces of evidence to demonstrate this, including the advantages of the so-called “fine-tuning”
approaches such as the Reinforcement Learning with Human Feedback (RLHF) method used in various LLMs.

%We argue that the growth and success of LLMs are to a great extent due to the fact that they are moved from pure AI algorithms to human-algorithm centaurs. We present various evidence to demonstrate this, particularly by focusing on the advantages of the so-called “fine-tuning” approaches such as the Reinforcement Learning with Human Feedback (RLHF) method used in various LLMs (e.g., OpenAI’s GPT-4, Anthropic’s Claude, Google’s Bard, and Meta’s LLaMA 2-Chat). 

We also discuss evidence showing that symbiotic learning on human intuition data can turn Generative AI tools into cognitive models capable of representing human behavior. We also discuss models in which only some of the conditions for creating centaurs are fulfilled, including In-Context Learning or Chain-of-Thought reasoning methods. 
We conclude by discussing two main points: (1) recent advancements in creating centaurs have moved us closer to reaching the goals that the founding fathers of AI—John McCarthy, Marvin Minsky, Nathaniel Rochester, and Claude Shannon—stated in 1955 as part of their proposed 2-month, 10-man study of AI to be held at Dartmouth; and (2) the future of AI development and use in many domains will most likely need to focus on centaurs as opposed to other AI approaches.

\section{Key Ideas}

While the concept of centaurs might sound similar to other hybrid paradigms facilitating collaboration between humans and machines such as Human-In-The-Loop-based (HITP) methods,  centaurs propose an inherently different Human-Machine Interaction (HMI)  paradigm that %views the relation between the human and the machine as a deep consolidation rather than a superficial cooperation. 
creates a \textit{merge} between humans and algorithms as opposed to just a simple
\textit{partnership.}
Similar to the fact that two businesses might either create a partnership or merge,
humans and algorithms can either work alongside each other through a simple partnership (while
being two separate entities that may or may not listen to one another) as suggested by HITP-type methods, or combine their powers
through a strong merge, as suggested by centaurs. 
%A helpful metaphor for the relation between the human and the machine in centaur-type models versus more traditional HMI schemes, such as HITP, is a business-model metaphor: Given two companies centaur may be perceived as a long-time fixed merger whereas traditional HCI schemes, on which we elaborate in Section ?, can be perceived as either short-term collaborative agreements, or a long-term acquisition in which the stronger organization takes over the weaker one rather than joining forces to give rise to a new entity as in a merger. 
The unique characteristics of centaurs that allow them to facilitate such strong merge-based collaboration is the use of \textit{symbiotic learning} and \textit{human intuition data.} That is, their strength is in both \textit{how} they learn, and \textit{what} they learn, and the existence of \textit{both} components is critical for creating centaurs. Section \ref{sec:symbiotic} provides an elaborated definition of centaurs and symbiotic learning, and presents a mathematical framework of centaurs.

Given that the two core components of centaurs are symbiotic learning and learning from human intuition data, a key question that should be posed is whether intuition is at all useful for learning meaningful generalizations.  
After all, intuition might seem the opposite of rigorous, or logical, or formal \citep{otte}. The consensus of most works is that while statistical rules indeed lead to more accurate results on unconstrained decision-making problems, intuition can surprisingly yield better results in some constrained decision-making problems, where typical constraints include incomplete data, uncertainty, and limited decision time, among others \citep{otte,fastslow,strategic,hbrnew}. In data-oriented terms, such constraints may refer to insufficient training instances or the lack of representative training cases, missing values for some of the features in the test case, and limited learning time or inference period. Centaurs, in essence, provide a strong merge that benefits from the {\em complementary} effect of intuition-based human decision-making and algorithmic decision-making \citep{sor1,sor}, providing a single entity that performs well in both constrained and unconstrained situations. Another scenario where human intuition, and specifically, training a model on human intuition data can be useful is when the loss function used prioritizes decisions that coincide with a human-like decision-making mechanism, even if that does not coincide with the optimal decision to take in the environment. We elaborate on specific cases in which the imitation of human decision-making can be beneficial in Section \ref{sec:discussion}.

Understanding that benefiting from intuition can indeed be useful, the next question is, when can it be trusted for making high-quality decisions?
Simon and Kahneman define intuition succinctly as unconscious pattern recognition \citep{simone,fastslow}. Kahneman describes the recognition-primed decision model as follows: When faced with a problem in a domain of expertise, we initially do not try to come up with several possible courses of action and then evaluate which one is the best. Instead, our associative memory automatically draws on a collection of patterns in our subconscious mind and comes up with the closest one \citep{fastslow}.  
Based on this definition, Hogarth \citep{hogarth}, followed by Kahneman and Klein \citep{kk,klein} suggested that intuitive decisions can only be trusted if they reflect a rich and repeated pattern-matching-like experience in environments with reliable feedback. That is, the environment should be stable enough so that past patterns can still be applicable to new ones. Furthermore, meaningful feedback should be constantly provided,  and a rich enough set of stored patterns as well as efficient pattern-matching methods should be readily available. 

Through symbiotic learning, centaurs ensure the validity of the use of intuition, as the above works suggest.  By choosing a model trained on a large set of patterns and with good-enough pattern recognition abilities, the continuous mutual learning process between the human and the machine provides meaningful feedback to both sides. And the fact that centaurs limit the extent to which intuition can influence the model’s weights, mean that they effectively utilize human intuition only to the extent that is beneficial.

The former becomes even stronger when the model used for pattern-matching is a generative model, including the modern architectures of LLMs as we thoroughly elaborate in Section \ref{sec:llm}. Traditionally, discriminative models were considered preferable to generative models on pattern recognition tasks \citep{generativebook}. While generative models were still considered preferable in cases where training sets are small, fast and accurate inference algorithms for complex generative models have been the bottleneck for the application of such models to large-scale instances \citep{generativebook}. 
A major step towards more effective inference in generative modeling has been the Transformer architecture \citep{transformer} and its application to LLLMs. Notably, recent works have demonstrated impressive results obtained by LLMs on a variety of pattern-recognition problems \citep{pattern,pattern2}. 
Apart from the high effectiveness of LLMs for pattern recognition tasks, an additional benefit that makes LLMs highly suitable to be used as centaurs is the use of methods such as Reinforcement Learning with Human Feedback (RLHF) \citep{rlhforiginal}. These methods allow reliable human-based feedback in the form of shaped rewards while guaranteeing the environment remains stable enough by constraining the differences between two consecutive policies.

While modern Generative AI constructs such as LLMs incorporate a built-in collaboration between humans and algorithms in the form of prompting and In-Context Learning, we note that it is unclear to what extent such collaborations have a true effect on model learning. In fact, existing literature \citep{prompt1,prompt2} suggests that prompting does not result in real learning, but rather functions merely as a sophisticated “query
normalization” method for building better encoding of user queries. That is in contrast to one of symbiotic learning’s inherent traits, the human input having a direct influence on the model’s weights. We stress that this is only one view of prompting, and understanding the exact influence of prompting on LLMs' learning and inference is still a matter of debate. 

Newer models, such as the newest release of OpenAI’s o1 \citep{o12} take one step forward and incorporate human-machine collaboration not only via ad-hoc prompting, but also as an inherent part of the model's architecture in the form of human-produced Chains-of-Thoughts (CoTs). While CoTs have been commonly used in an In-Context Learning setting for enriching prompts (CoT Prompting), o1 incorporates such human-produced CoTs within its learning and reasoning process using Reinforcement Learning (RL), where the overall objective is for the model to understand not only what a good outcome is, but also what a good ``reasoning process" is.  
The incorporation of human-produced CoTs within the model's learning and reasoning process can certainly be seen as one step forward toward creating centaur models; however, we argue that such models are still different than centaurs and, hence, do not provide the strong human-algorithm merge we introduced earlier. The key difference lies in the definition of intuition as suggested by \citep{simone,fastslow}, and specifically on its inherent trait of being  “unconscious.”  According to \citep{klein}, this fact can be attributed to the reliance of intuition on {\em tacit knowledge}: knowledge gained through experience. Tacit knowledge {lacks structured order, is difficult to explain, and often unavailable to consciousness} \citep{klein,fastslow,simone}. CoT prompting or learning represents the opposite: it is a {structured, conscious, and purposely explainable} human input that aims at conveying purely logical arguments. Models trained on CoT human input alone, thus, might not be considered full centaurs, since they lack human intuition.

In closing this section, we note that our aim
in this paper is mainly  (a) starting to raise important questions about human-algorithm centaurs, and (b) providing the first steps for encouraging a wide-ranging debate about their potential in making future AI models more effective. The definitions and arguments we provide, thus, are not black-and-white. Furthermore, they might change over time, especially since it is up to future scholars to create highly effective centaur-based AI models.

The rest of this document is organized as follows. Section \ref{sec:motivation} presents the motivation for creating and using centaurs. Section \ref{sec:loop} characterizes common existing methods for human-machine collaboration, clarifying why centaurs are different.  Section \ref{sec:symbiotic} presents the notion of symbiotic learning and provides a concrete mathematical framework for creating centaurs. Section \ref{sec:llm} discusses why  LLMs are highly suitable as base models for creating centaurs, and presents evidence suggesting that centaur-based LLMs can function, at least partially, as cognitive models. Section \ref{sec:discussion} discusses the advantages and disadvantages of centaur models, their potential use cases, and some ethical considerations. Finally, Section \ref{sec:conclusion} concludes by discussing why future use of AI in various applications will likely be centaur-based.

\section{Motivation}\label{sec:motivation}

%“The intuitive mind is a sacred gift and the rational mind is a faithful servant. We have created a society that honors the servant and has forgotten the gift.”  (\citep{einstein})

Humans often face complex decisions, and it seems that their intuition is not always helpful. When facing critical life-changing decisions such as quitting a job or ending a relationship, we tend to be happier with the outcomes later on, when a coin toss tells us to make a change than when it promotes maintaining the status quo \citep{9}.

Nonetheless, human intuition is often very powerful, especially when we want to make quick decisions. Put it differently, while intuition often misfires when we are dealing with complex problems that require careful analytics (e.g., in predicting whether a particular startup will be successful and is worth large investment \citep{startup}, finding ways to reduce incidents of diabetes for organ transplanted patients \citep{10,11}, deciding upon cell formation and layout design for a cellular manufacturing system \citep{12},  or finding most effective ways of saving lives in emergency rooms \citep{sor2,13,14}), it can be very useful when using data, models, and careful analytics is not an option. Malcolm Gladwell’s popular book “Blink: The Power of Thinking Without Thinking” provides various examples of this, including when police officers need to quickly decide whether to shoot a suspect \citep{15}. Good intuitive decision-making also helps fire fighters when they face a burning building \citep{16}.

A central question, however, that might arise is this: when and why should one incorporate human intuition in AI-based decision making processes? Wouldn’t it degrade the overall performance? After all, aren’t AI models better than human intuition in many tasks? 

While relying on intuition in handling complex problems can be misleading, combining intuition with the most useful analytics approaches can often be better than just relying on analytics.  To better understand this, it is useful to see how our own system of thinking works. Daniel Kahneman---the famed psychologist, known for his groundbreaking work on the psychology of judgment and decision-making as well as behavioral economics, who won the Nobel Prize in Economics in 2002—highlighted in his book “Thinking Fast and Slow” that our brain has two modes of thinking: System 1 and System 2. System 1 is fast and instinctive, but System 2 is slower, more deliberative, and more logical. What is perhaps more interesting is that these systems greatly complement each other. Our body somehow knows that we need both systems to be able to make good decisions in different situations. 

Similar to how System 1 and 2 complement each other, intuition and analytics can help each other as well. And this is where human-machine centaurs can play a vital role. We—humans—can use our intuition in many ways while developing analytics methods and taking advantage of computers to run them. For starters, intuition often allows us to develop better models, or considering the George Box’s aphorism “all models are wrong, but some models are useful,” more “useful” ones. Intuition also allows us to verify the results obtained from models and make sure that the assumptions made in the model are not problematic. Analytics scientists often use this simple technique as a feedback loop: when the results obtained from a model are not sensible and can be related to a wrong assumption in the model, they modify their model to obtain a better one.  And if they are working with a cloud of models to address the curse of ambiguity \citep{17,18}, they can replace the models that might be causing preposterous results. Preposterous results could also be related to abnormalities and/or outliers in data that need to be removed before feeding them to models, and intuition is often very helpful here as well.

 The complementary functionalities of formal analytics and human intuition are the key to centaurs' potential of working better than the best algorithms, even in tasks in which human intuition is relatively weak. In the first author's experiments with the Mayo Clinic, for example, the ranking in performance (from best to worst) was as follows: human-algorithm centaur, algorithm, and human experts. Notably, the best human expert’s performance was significantly below that of an AI model that was not benefiting from human intuition. Yet, feeding human intuition to it could create a desirable boost in performance \citep{sor, sor1}.

Realizing that intuition alone can be misleading in understanding and analyzing complex systems, and that human-machine integration is needed to harness the full power of both advanced analytics and human intuition, has introduced new methods of creating human-algorithm centaurs capable of symbiotic learning, as will be discussed in Section \ref{sec:symbiotic}.
%We saw some of such methods in the previous section (Figure 1). In Section 4, we will elaborate more on specific methods that Generative AI models such as recent LLMs have used (Figure 2). However, prior to doing so, it is useful to first see some of the main advantages of centaurs vis-à-vis traditional ML and AI models.

\vspace{10pt}
\noindent
\subsection{Main Advantages of Centaurs Compared to Other AI Models}
%Beyond the fact that human-algorithm centaurs can outperform both best algorithms and best human experts, 
There are various reasons why developing and implementing centaurs should be on top of mind in different sectors: 

1. Enhanced interpretability: While some models such as decision trees are inherently interpretable, other common models such as neural networks lack interpretability. Incorporating human intuition into the learning process of models may allow humans to better understand their predictions as well as the inference process that led to those predictions, thus increasing the interpretability level of naturally uninterpretable models. This observation was empirically validated by \citep{prior}, which shows that explicitly including human feedback in the algorithm's learning process improves the algorithm's interpretability on some datasets even to a larger extent than methods imposing direct interpretability constraints.

2. Reduced algorithm aversion:  Algorithm aversion has been widely reported as a main hindrance in creating impactful AI in applications in which the final decision-maker is a human \citep{31, 32, 33}.  Through experiments, we have found that utilizing centaurs can reduce algorithm aversion, mainly because recommendations from centaurs are closer to human way of thinking since centaurs incorporate human intuition \citep{sor}. For example, using the “weight on advice measure,” we found that physicians are more willing to consider recommendations that come from centaurs than those from pure AI models \citep{sor}. Furthermore , the recommendations coming out of centaurs represent less “human aversion” in that, not surprisingly, they better represent human behavior and preferences.  

3. Reduced causation aversion: Centaurs can help AI developers overcome what we might call “causation aversion.” This refers to the fact that most of the focus of the current AI algorithms has been on the association layer of the “Ladder of Causation” \citep{41}. What decision-makers in various sectors need are algorithms that can work beyond the association between various input and output variables. Prescriptive analytics, where the intention is to recommend decisions that can cause improvements, can benefit a lot from moving algorithms to human-algorithm centaurs. Given various ongoing research on causal ML (see, e.g., \citep{10} and the references therein), one can expect to see more efforts in developing centaurs capable of causal and counterfactual reasoning.

4. Enhanced adaptability to behavioral tasks: Centaurs achieve better results on behavioral tasks---tasks that require the model's output to coincide with human behaviors such as human-study replication \citep{studies3}. Likewise, centaurs show better results on tasks evaluated using behavioral evaluation metrics \citep{intentions,firstllmreward2}. For instance, chatbots are often evaluated according to the extent to which human users found them ``helpful''.
We further elaborate on the use of centaurs for behavioral tasks in Section \ref{sec:discussion}.

5. Better specification of hard-to-define objectives: Traditionally, AI models have been applied to tasks
for which it is quite easy to come up with well-specified objective functions. However, as these models have been widely adopted in numerous areas and for a wide array of diverse tasks beyond the classification of well-structured data, many of such newer, more complex tasks involve goals that are hard to define, and thus it is not immediately clear how to design a cost function for such tasks. 
Instead of explicitly defining an objective function for such tasks, the objective can be dynamically learned using human intuition in the form of preferences or demonstrations. The model then follows the human-guided, learned objective function \citep{rlhforiginal,demons}. 

6. Enhanced performance on low-quality datasets: A key trait of AI models is the ability to generalize from one dataset, a training set, to other unseen datasets. However, training datasets may contain flaws that inhibit the model's generalization ability. For instance, the training set at hand can be unrepresentative of the general population. Alternatively, the training dataset may be representative of the general population at \textit{train time}, $t_{train}$, but not at \textit{deployment time}, $t_{test}=t_{train}+t'$, (\textit{i.e.} concept drift). Human intuition can be used to mitigate such generalization caveats by adding more nuanced yet general reasoning into the model, enabling it to generalize to unseen, potentially novel, examples \citep{conc1,conc2,conc3,conc4,conc5}.

\section{Centaur Versus Human-in-the-Loop Approaches}\label{sec:loop}

In 1960, J.C.R. Licklider published a paper entitled “Man-Computer Symbiosis” in which he popularized the idea of man-computer symbiosis “as an expected development in cooperative interaction between men and electronic computers” \citep{mancomputer}. In his words, the aim was to “enable men and computers to cooperate in making decisions and controlling complex situations without inflexible dependence on predetermined programs.” In today’s analytics science, we can think of symbiotic learning and the incorporation of human intuition as two important characteristics of centaurs. These two characteristics distinguish centaurs from other related designs in AI such as those under the ``human-in-the-loop'' umbrella. We discussed the advantages of incorporating human intuition in the previous section. In the next section, we will discuss the other distinctive feature of centaurs: symbiotic learning. Here, we start by reviewing human-in-the-loop approaches so as to distinguish them from centaurs.

\textbf{Active-learning approaches.} Some works (see, e.g., \citep{active1}) model the human as an \textit{oracle}: given a query on a subset of unlabeled data records, issued by the AI model, the human responds with the labels of the data records in the query. The human in this case is used as a ``data annotator.'' After the query's response is submitted back to the AI model, the annotated records are used to improve the model's performance by conducting either full or partial retraining of the model using the annotated records. Active learning approaches are particularly useful when the vast majority of the available data is unlabeled and the cost of annotations is high. Thus, it is a popular approach in domains such as healthcare and finance \citep{financeact1,financeact2,healthact1,healthact2}. 

\textbf{Interactive-learning approaches.} Some other works (see, e.g., \citep{incfirst}) also model the human as an oracle, but this time the human itself can actively control \textit{which} queries are submitted to her and on which ones she will respond with annotations. Unlike an active-learning process which may or may not be iterative, interactive-learning algorithms are both iterative and incremental in nature \citep{inc1,inc2}; the human is actively participating not only in the annotation process but also in the query creation process and/or in the model evaluation process aimed at assessing whether more queries are needed \citep{inc4,inc5}.

\textbf{Workload partitioning approaches.} In this set of approaches, one typically divides the prediction workload between the human and the machine based on pre-selected criteria. That is, each record is processed either by the machine, the human, or both the machine and the human where the goal is to minimize the number of records that humans are required to process. This approach is oftentimes used in domains in which, on the one hand, experts' annotation time is expensive, yet, on the other hand, the degree of error tolerance is low. In such case, the minority of ``complex'' cases will be processed by the human whereas the majority of cases will be processed by the machine. The allocation of records either to the human expert or to the model can be done either based on manually selected criteria such as the risk associated with the record, or based on automatic criteria in which the model itself can decide who should be assigned with the record (e.g., based on its confidence level on the record). Workload partitioning schemes are thus oftentimes deployed in high-risk domains, such as applications related to security and health. While in some works humans and machines share the same workload \citep{work3,work4,refprofasked}, in most workload-partitioning works the majority of the workload is put on the AI algorithm, and only the minority of harder or riskier tasks are assigned to humans \citep{work1,work2}.

\textbf{Machine teaching approaches.} The main human functionality in machine-teaching schemes is typically in designing an optimal training set for the AI algorithm \citep{teachingf1,teachingf2}. Importantly, machine-teaching approaches often assume that the human already knows the ``true'' decision boundary, and its goal is merely to convey that boundary to the model via the selection of a proper training set. Of course, the human can also use its control on the training set to guide the model towards a different decision boundary than the true one. Indeed, most applications of the machine-teaching approach are either in the security domain, which uses machine teaching to construct both attacks and defense schemes on automated learning agents \citep{secteach1,secteach2} or in the education domain, which uses machine teaching to build models that are better at conveying information to human learners \citep{eduteach1,eduteach2}.

\textbf{Curriculum learning approaches.} These set of approaches are similar to machine teaching approaches in that the main human functionality is organizing the training set used by the AI model \citep{cloriginal}. Unlike machine teaching, the human in curriculum-learning schemes is not assumed to know the true decision boundary. Instead, curriculum learning assumes the human to have a close familiarity with the prediction task at hand. Using that familiarity, and given a fixed set of training data, the human's goal is designing an optimal curriculum based on the fixed training data. The idea behind curriculum learning is that the learning process can be quicker if the examples are not randomly presented but rather organized in a meaningful order, where the common wisdom is that ``easier'' examples should be presented before ``harder'' examples \citep{cloriginal}. Defining a proper ``difficulty'' measure is thus the main challenge when designing a curriculum-learning scheme, and different works use different difficulty measures. Common difficulty metrics in the Natural Language Processing (NLP) 
 domain include sentence length \citep{cl1}, parse tree depth \citep{cl2}, and level of word rarity \citep{cl3}. Common difficulty metrics in the vision domain include number of objects \citep{cl4}; shape variability \citep{cloriginal}; and inter-image clustering density \citep{cl5}.

\textbf{Meta-Learning approaches.} In meta-learning approaches, human feedback is neither used for preprocessing actions such as training set organization nor is combined in the learning process. Instead, human feedback is used for different meta-learning tasks such as model evaluation, hyperparameter tuning, and model calibration, among others. Importantly, such feedback is not used to further train the model, but merely to further improve the model without directly combining the feedback in the training process. For example, \citep{anomalyensemble} builds an anomaly-detection ensemble and uses human experts' opinions on a subset of query records for tuning the ensemble's hyperparameters: the weight assigned to each model in the ensemble. Likewise,  \citep{calibration} presents a human-preference-based calibration approach, suggesting calibration with respect to the human’s confidence in her own
predictions produces calibrated estimates that lead to optimal decisions. Finally, \citep{humaneval1,humaneval2} use humans as model evaluators, where humans observe the learned model's outputs on a representative set of inputs and evaluate its output quality by ranking to what extent it aligns with their subjective preferences.

\textbf{Other approaches.} {More recent methods for human-machine collaboration such as Human-Guided Rewards, Human-based Prompt Engineering, Chain-of-Thought Reasoning, Supervised Fine-Tuning on Human Intuition Data, and more, are discussed in Section \ref{sec:symbiotic} and Section \ref{sec:llm}.}

In almost all of the approaches discussed above, the human functions mainly as a ``service provider:'' she is either treated as an oracle, providing labeling services, or as a teacher, providing preprocessing services such as training set design. {Notably, the human is not treated as an \textit{equal} of the learning model who can directly incorporate her own intuition---in the observed form of preferences, actions, decisions, or opinions---as part of the learning process. 
Such observed signals of human intuition result from the human's unique judgment, which is subjective in nature, as opposed to the human's perception as an oracle where the human input is assumed to always coincide with a certain objective truth (e.g. ground-truth labels).} In those human-in-the-loop approaches where humans are not merely used as service providers but are rather queried for their own subjective input, such inputs are not combined directly in the learning algorithm. Instead, they are used \textit{orthogonally} to the AI model, either by working on a different set of records than the machine (workload partitioning approaches) or by providing subjective feedback that is not incorporated directly in the learning process (meta-learning approaches). 

As we will discuss in the next section, symbiotic learning which is a core element of centaurs makes them different from these human-in-the-loop approaches by taking a broader approach to the cooperation between humans and machines, treating both parties as equals rather than a-priori giving the lead to one party. Such a holistic approach resembles the concept of Human-Intelligence Collectives (HIC) \citep{hic,hic2}, which researches the creation of human-agent teams and the best cooperating and control methods for achieving a predefined set of goals that may defer among team members. Importantly, HIC considers each member of the team to be an autonomous stakeholder, having its own set of goals which may not coincide with other members' goal. Thus, HIC mainly focuses on the research of new incentive engineering and reward modeling techniques, rather than on the research of mutual human-AI learning methods under commonly shared goals as is the case with centaurs.

\section{Symbiotic Learning:  A Core Characteristic of Centaurs 
}\label{sec:symbiotic}

As mentioned in previous sections, the distinguishing characteristic of centaurs from human-in-the-loop approaches is symbiotic learning.
Symbiotic learning departs from the methods discussed in Section \ref{sec:loop} in that it implements a mutual human-AI learning process where the human is not treated merely as an oracle or a preprocessor, but rather as an equal partner in the learning process by incorporating her unique input into the learning process. {Human input includes observed signals of human intuition such as preferences, behaviors, or opinions, and unlike ground-truth labels, such inputs are inherently subjective. The core differences between human-in-the-loop approaches as described in Section \ref{sec:loop} and symbiotic learning are three-fold:}

%1. Symbiotic learning allows for the fact that the human input is often in the form of intuition, a subjective opinion, a behavior, or a preference rather than a ground-truth label.
{1. Symbiotic learning facilitates multi-input learning, in which learning is performed using multiple datasets corresponding to either the same set of observations or different sets of observations.  At least one dataset represents human input, and includes observed signals of human intuition such as subjective preferences, behaviors, opinions, or decisions.}

2. The human input is directly incorporated in the learning process of the downstream model's parameters (e.g., weights), which we denote by $\theta_{symbiotic}$.

{3. The influence of human input data on the final model's parameters is constrained to ensure appropriate reliance on human intuition given the specific traits of the environment, the degree of human expertise, and feedback quality. The constraints are optimized either as a separate learning task or jointly with $\theta_{symbiotic}$.}

\begin{figure}[tb]
\includegraphics[height=1.4in, width=5.4in]{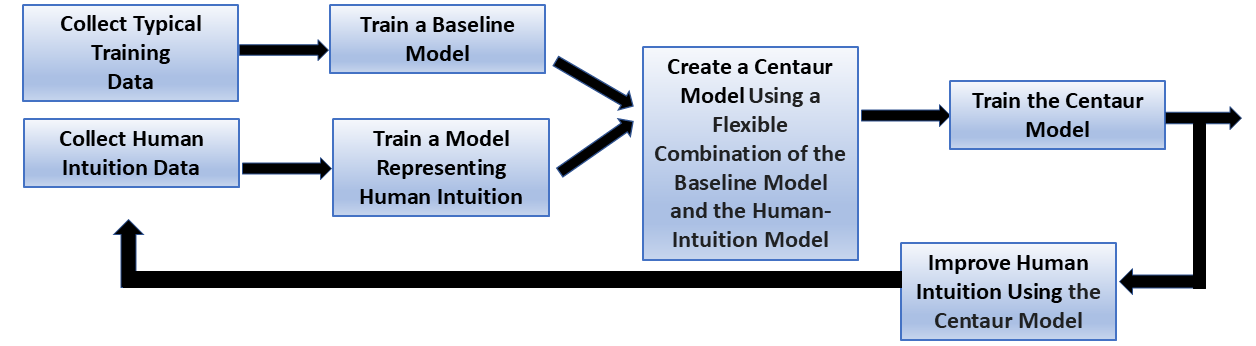}
\caption{ \small A high-level representation of symbiotic learning}
\label{fig:General}
\end{figure}

Symbiotic learning defines a new paradigm, where humans do not function merely as service providers supporting the AI model, but rather as full partners to the model in the learning process. To account for a wide array of tasks and applications, the nature of such partnership is learned on a case-by-case basis using flexible constraints. A high-level diagram of symbiotic learning is given in Figure \ref{fig:General}. In what follows, we present a mathematical framework that further details this representation.

We frame symbiotic learning as a general constrained optimization problem, abstracting away specific symbiotic learning techniques using different loss functions, models, and constraints. A parametric example of our symbiotic learning framework is:

\begin{equation}
\label{eqn:sym1}
 \hat{\theta}_M = \argmin_{\theta \in \Theta} 
\mathbb{E}_{z \sim d_{data}} l_{data}(z;\theta)
\end{equation}
\begin{equation}
\label{eqn:sym2}
\hat{\theta}_H = \argmin_{\theta \in \Theta} 
\mathbb{E}_{z \sim d_{human}} l_{human}(z;\theta)
\end{equation}
\begin{equation}
\label{eqn:sym3}
\hat{\theta}_{symbiotic}=\gamma_{symbiotic}(\hat{\theta}_M,~\hat{\theta}_H) = 
\argmin_{\gamma \in \Gamma,~\mathcal{C}(\gamma, \hat{\theta}_M) \leq c_1,~\mathcal{C}(\gamma,\hat{\theta}_H) \leq c_2}
\mathbb{E}_{z \sim d_{human}, d_{data}}[l_{comb}(z;\gamma;\hat{\theta}_M;\hat{\theta}_H)] 
\end{equation}
\begin{equation}
\label{eqn:sym4}
h_{\theta_M}: \mathcal{X}_M \rightarrow \mathcal{Y}_M,~~~~
h_{\theta_H}: \mathcal{X}_H \rightarrow \mathcal{Y}_H,~~~~
h_{\theta_{symbiotic}}: \mathcal{X}_S \rightarrow \mathcal{Y}_S
\end{equation}

Given a data distribution $p_{data}$ and a corresponding empirical sample $d_{data}$, we train a base model parametrized by $\hat{\theta}_M$ to minimize an arbitrary loss function, $l_{data}$, on $d_{data}$. Given a human-preference distribution $p_{human}$ and a corresponding empirical sample $d_{human}$, we train a human preference model parametrized by $\hat{\theta}_H$ to minimize an arbitrary loss function, $l_{human}$, on $d_{human}$.  
We
denote the space of symbiotic-model parameters by $\Gamma$, and model symbiotic learning as minimizing an arbitrary combination loss $l_{comb}$ on either $d_{data}$ or $d_{human}$ w.r.t
symbiotic learning parameters $\gamma \in \Gamma$ and subject to  constraints,  $\mathcal{C}$, on either $\hat{\theta}_M$, $\hat{\theta}_H$, or both. Finally, the learned parameters can be used to learn the functions denoted by $h$ in the last line, where a mapping between the space of inputs to that of outputs occurs. In this setting, loss functions are quite general and include various aspects such as regularizations that is often used in high-dimensional settings to avoid over-fitting.

In practice, different symbiotic learning techniques use different instantiations of the definition in Equations \eqref{eqn:sym1}-\eqref{eqn:sym4}.
For instance, in human-preference-based supervised fine-tuning approaches no $\theta_H$ is trained, and thus $l_{comb}$ is optimized solely on $d_{human}$ given constraints on $\hat{\theta}_M$. In preference-based augmented-covariate-space approaches, no $\theta_M$ is trained and thus $l_{comb}$ is solely optimized on $d_{data}$ given constraints on $\hat{\theta}_H$. In preference-constrained cost function approaches, neither $\theta_M$ nor $\theta_H$ are trained, and thus the symbiotic learning framework collapses into a simple constrained loss function using a tunable hyperparameter that controls the importance given to human preferences within the loss. In human-machine ensemble approaches, both $\theta_H$ and $\theta_M$ are trained, and $l_{comb}$ is optimized on $d_{data}$.

Distributions $d_{data}$ and $d_{human}$ can also take different forms in different symbiotic learning techniques. Generally speaking, in supervised learning applications of symbiotic learning such as the one discussed above we assume $d_{data}$ to contain pairs of the following form: 
$z_i=(x_i,y_i) \in \mathcal{X}_{data} \times \mathcal{Y}_{data}$
and $d_{human}$ to contain pairs of the following form:
$z_i = (x^h_i, y^h_i) \in \mathcal{X}_{human} \times \mathcal{Y}_{human}$.
 In some symbiotic learning techniques, both $d_{data}$ and $d_{human}$ must represent the same underlying distribution, while in other techniques they may represent two different distributions. For instance, an underlying assumption of preference-based augmented-covariate-space approaches (see below) is that $d_{data}$ and $d_{human}$ represent the same underlying distribution, where $\mathcal{X}_{data}=\mathcal{X}_{human}$, $\mathcal{Y}_{data}=\mathcal{Y}_{human}$, whereas in human-preference-based supervised fine-tuning approaches (see below) $d_{data}$ and $d_{human}$ can take different distributions. In fact, in the latter case $d_{human}$ is often labeled, where $y^h_i$ corresponds to human feedback, whereas $d_{data}$ is unlabeled, and thus $\mathcal{Y}_{data}=\emptyset$.

\vspace{10pt}
We now discuss multiple concrete techniques for the implementation of symbiotic learning. These techniques are
specific ways of utilizing the general framework presented via Equations \eqref{eqn:sym1}-\eqref{eqn:sym4}.

\vspace{5pt}
\textbf{Preference-based augmented covariate space.} In this approach, $\gamma$ refers to the learned parameters (e.g., weights)
 of a model, $h_{\gamma}$: $\mathcal{X}_S \rightarrow \mathcal{Y}_S$, where $\mathcal{X}_S$ is an augmented version of $\mathcal{X}_{data}$ which contains for each record, $i$, its corresponding human feedback $y^h_i$ extracted from $d_{human}$ so that $h_{\gamma}$ is fed, for each record, $i \in d_{data}$, the following pair: $(x_i \cup \{y^h_i\},y_i)$. The new set of features for a record, $i$, contains its original set of features, $x_i$, as well as the human feedback on that record, $y^h_i$.

A practical caveat of that approach is that human feedback is needed on each record in $d_{data}$, both in training and in production, and hence the following must hold $|d_{human}| \geq |d_{data}|$, which is impractical in most scenarios. 
An alternative approach is to collect human preferences on a different, smaller dataset, so that $|d_{human}| \ll |d_{data}|$ while $d_{human}$ is highly representative of $d_{data}$. To this end, for
 for each $i \in d_{data}$ one can search for its k-nearest neighbors in $d_{human}$, and use a combination of their assigned feedback ($y^h_i$), such as a majority vote, as an additional feature for $i$ in $h_\gamma$. 
However, it is not immediately clear how to choose the best distance metric between samples in $d_{data}$ and samples in $d_{human}$, and the results might be sensitive to the choice made for this metric.

A generalization of the nearest-neighbor approach presented in \citep{sor} is to implicitly learn such similarity relations by building a human-preference model and use its prediction as an additional feature to $h_\gamma$, as opposed to using the raw preferences $y^h_i$. 
That is, one can first learn a human preference model $h_{\theta_H}$ using $d_{human}$:

\begin{equation}  
\hat{\theta}_H = \argmin_{\theta \in \Theta} 
\mathbb{E}_{z\sim d_{human}} l_{human}(z;\theta).
\end{equation}
For each $i \in d_{data}$, one can then use the value $h_{\theta_H}(x_i)$ as an additional feature for learning a model 
$h_\gamma$, trained to minimize the loss over $d_{data}$. In this setting, the set of constraints would be 
$\mathcal{C}(\gamma, h_{\theta_H})$ which denotes constraints on the importance assigned to feature $h_{\theta_H}(x_i)$ during the learning process compared to other features in $\mathcal{X}_{data}$.

\vspace{5pt}
 \textbf{Human-guided rewards.} As in the preference-based augmented covariate space approach, this approach starts by
 learning a human preference model, $h_{\theta_H}$, using the human preference data, $d_{human}$:
\begin{equation}
\hat{\theta}_H = \argmin_{\theta \in \Theta} 
\mathbb{E}_{z\sim d_{human}} l_{human}(z;\theta).
\end{equation}
One then uses the outputs of $\hat{\theta}_H$ on samples in $d_{data}$ for estimating a newly-learned symbiotic learning model, $h_\gamma$ with newly-learned parameters, $\gamma$.
However, unlike the preference-based augmented covariate space approach, the outputs of $h_{{\theta}_H}$ are not embedded within $h_\gamma$'s feature space, $\mathcal{X}_S$, as additional features, but rather used either consistently or sporadically to guide $h_\gamma$ towards solution regions compatible with the human input in $d_{data}$. Moreover, unlike prior symbiotic learning schemes where the exact form of $\hat{\theta}_H$, as well $l_{human}$ used for training it was rather flexible, here the human preference model, $\hat{\theta}_H$, also referred to as the reward model, must have a specific structure. First, a reward model must take as input pairs $(x,y)$ where $x$ denotes a context and $y$ denotes a possible output. Second, the output of the reward model must be a scalar, denoting the predicted human-preference score given to output $y$ created in the context of an input $x$. Thus, $\mathcal{X}_H=\mathcal{X}_S \times \mathcal{Y}_S$
 and $\mathcal{Y}_H=\mathbb{R}$.
 $\hat{\theta}_H$ is trained to maximize the rewards of outputs that are highly preferred by humans while minimizing the reward of outputs that are less preferred by humans. Specifically, $l_{human}$ often takes the following general form:
\begin{equation}
l_{human}(\hat{\theta}_H)=\mathbb{E}_{(x,y_1,y_2)\sim d_{human}}[f_1(f_2(\hat{\theta}_H(x,y_1)-\hat{\theta}_H(x,y_2)))],
\end{equation}
where the training set $d_{human}$ is composed of triplets $(x,y_1,y_2)$: $x$ denotes a context (an input to $h_\gamma$), $y_1$ denotes the output that is more preferred by humans in the context of $x$, and $y_2$ denotes the output that is less preferred by humans in the context of $x$. $f_1,f_2$ are task-specific and may correspond, for instance, to the $log()$ and $sigmoid()$ functions, respectively.

In some configurations, $\hat{\theta}_H$ is learnt first, and then used for learning $h_\gamma$. In other configurations, both $\hat{\theta}_H$ and $h_\gamma$ are jointly learned, where the output of $h_{{\theta}_H}$ is used as rewards to guide $h_\gamma$'s learning process and optimizing $l_{data}$ while at the same time outputs of $h_\gamma$ are used to further tune $\hat{\theta}_H$ using $l_{human}$. While human-guided rewards can be used for aligning different types of AI schemes, its most common use is for replacing or augmenting the reward function in reinforcement learning, creating Reinforcement Learning with Human Feedback (RLHF) schemes \citep{rlhforiginal, intentions}. As we will discuss later, RLHF in turn has played an important role in moving various generative AI models, including a vast majority of LLMs, from a pure AI model to a centaur-based system by incorporating human intuition and symbiotic learning---the two distinguishing characteristics of centaurs.

 Using human-guided rewards as opposed to a fixed reward function is especially useful for optimizing tasks in which it is easier to recognize the desired behaviors than explicitly defining them. In an RLHF-based setting, $l_{comb}$, which denotes the policy's loss, takes the following general form:
%\begin{equation}
%l_{comb}(\gamma)=\mathbb{E}_{(x,y) \sim D_{h_\gamma}} [\hat{\theta}_H(x,y)-\mathcal{C}(h_\gamma(y|x))],
%\end{equation}
\begin{equation}
l_{comb}(\gamma)=\mathbb{E}_{(x,y) \sim D_{h_\gamma}} [\hat{\theta}_H(x,y)-\mathcal{C}(h_\gamma(y|x),h_{\hat{\theta}_I}(y|x))],
\end{equation}
where $x \in \mathcal{X}_{data}$, $y=h_\gamma(x)$, and $\mathcal{C}$ denotes a constraint applied to the symbiotic model, aimed at preventing it from ``overfitting'' to human preferences by limiting its divergence from an initialization model, $\hat{\theta}_I$. We will elaborate on the use of human-guided rewards for LLM training later in this section where we use GPT-4 training as a specific example.

\vspace{5pt}
\textbf{Human-preference-based supervised fine-tuning.}  While the term ``fine-tuning'' is nowadays mostly used in the context of LLMs, it is more broadly a re-branding attempt of a collection of well-known, well-established transfer-learning-based methods \citep{transfer}. Nonetheless, some newer techniques are designed specifically for applications in LLMs as we will discuss in Section \ref{sec:llm}. A typical supervised fine-tuning task consists of taking an existing model trained on a given labeled or unlabeled dataset, and adapting it to another task using another labeled dataset.
Fine-tuning a model with human feedback can, thus, be thought of as first training a model $h_{{\theta}_M}$ using $d_{data}$: 
\begin{equation}
\hat{\theta}_M = \argmin_{\theta \in \Theta} 
\mathbb{E}_{z \sim d_{data}} l_{data}(z;\theta),
\end{equation}

and then adapting $h_{\theta_M}$ to human preferences using another dataset, $d_{human}$: 
\begin{equation}
\hat{\theta}_{symbiotic}=
\argmin_{\gamma \in \Gamma, \mathcal{C}(\gamma, \hat{\theta}_M) \leq c_1}
\mathbb{E}_{z \sim d_{human}}[l_{human}(z;\gamma;\hat{\theta}_M)].
\end{equation}

In the supervised case, $\gamma$ corresponds to a small set of parameters \textit{within} the same model $h_{\theta_M}$ so that most parameters are frozen using the initialization of $\theta_M$ and only few parameters, $\gamma$, can change during fine-tuning using $d_{human}$ subject to the constraints enforced by $\mathcal{C}$. For example, when using adapter layers \citep{adapter1,adapter2}, due to the high training cost only parameters corresponding to the last layers will be tuned while all other parameters are frozen according to $\theta_M$'s initialization. Details on supervised fine-tuning can be found in Section \ref{sec:llm}.

\begin{figure}[tb]
\includegraphics[height=3in, width=5.7in]{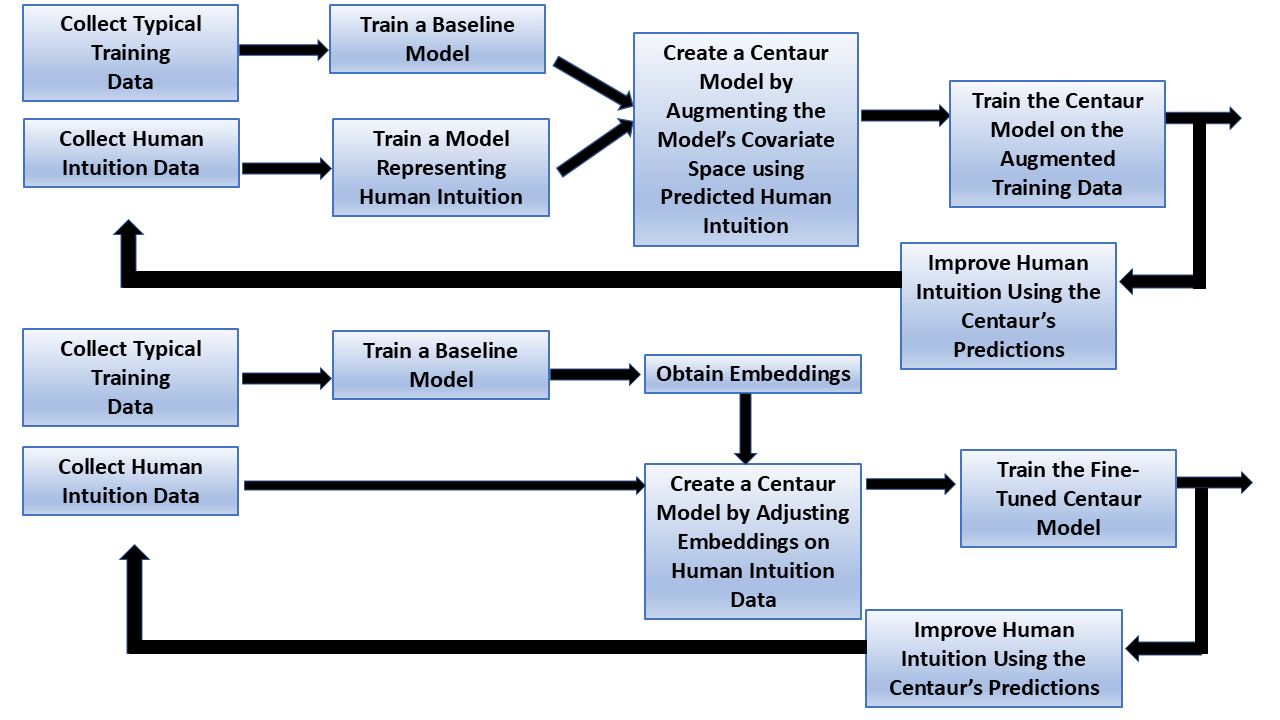}
\caption{ \small Two modern methods of creating centaurs. Top: Preference-based augmented covariate space. Bottom: Human-preference-based supervised fine-tuning}
\label{fig:Specific}
\end{figure}

\vspace{5pt}
\textbf{Human-machine ensembles.} Ensemble-based approaches view both the human and the machine as experts and aim at optimizing a joint model of the two types of experts \citep{ensemble1,ensemble2}. Specifically, we are given $N$ heterogeneous machine learning models, indexed by $n=1,2, \cdots, N$ and trained using
 $d_{data}$:
\begin{equation}
\hat{\theta}^n_M = \argmin_{\theta \in \Theta_n} 
\mathbb{E}_{z \sim d_{data}} l_{data}(z;\theta),
\end{equation}
as well as $K$ heterogeneous human-preference models indexed by $k=1,2, \cdots, K$  and trained
 using $d_{human}$:
\begin{equation}
\hat{\theta}^k_H = \argmin_{\theta \in \Theta_k} 
\mathbb{E}_{z \sim d_{human}} l_{human}(z;\theta).
\end{equation}

The goal is to construct an ensemble model 
$h(\hat{\theta}^1_M, \cdots ,\hat{\theta}^N_M,\hat{\theta}^1_H, \cdots ,\hat{\theta}^K_H)$ such that $h$ effectively combines $\hat{\theta}^n_{M_{n \in 1, \ldots, N}}$ and $\hat{\theta}^k_{H_{k \in 1, \ldots, k}}$ where $h$ is either a fixed combination function or a learnable model $h_{\gamma}$ with learnable weights, $\gamma$.

In the simplest form, $h_\gamma$ functions as a meta-learner using a stacking approach that learns the optimal weights of $\hat{\theta}_M$ and $\hat{\theta}_H$ in the final prediction, or further learns how to combine $\hat{\theta}_M$'s and $\hat{\theta}_H$'s predictions. In more complex configurations, the contribution of the human-preference model to $h_\gamma$ is further constrained.  An example of such a constrained ensemble setting is the GPT-4 architecture \citep{openai}, in which $\hat{\theta}_M$ corresponds to a pre-trained language model, $\hat{\theta}_H$ corresponds to a learned human-based
reward model, and $\gamma$ are the parameters of a fine-tuned model, $h_\gamma$. In GPT-4, $\gamma$ is
 learned using the pre-trained language model and the reward model while constraining the current policy, resulting from rewards provided by $\hat{\theta}_H$, not to diverge too much from $\hat{\theta}_M$. This later step of constraining the current policy is operationalized using 
a KL-divergence penalty applied on a per-token basis between $h_\gamma$ and $\hat{\theta}_M$.

Another example of such a constrained ensemble setting is reward-model ensembles \citep{rewardensemble}, an emergent solution to the challenge of reward over-optimization often seen on out-of-distribution datasets. In such a setting, there exists $K$ reward models, representing diverse sources of human intuition-based feedback. Each reward model starts from a pre-trained initialized model on $d_{data}$, $\hat{\theta}_M$, and then fine-tunes it on its own $d_{human}$ dataset to a different extent, which allows capturing various degrees of tradeoff between machine-based and human-based intuition. Here, $\gamma$ is the parameters of each model's Low Rank (LORA) adapters \citep{lora} which effectively combine all the fine-tuned reward models $\{\hat{\theta}^k_H(\hat{\theta}_M)\}_{k=1,\ldots , K}$ 
%(depending on pre-trained models) 
into a single ensemble model, $h_\gamma$.

\vspace{5pt}
\textbf{Preference-constrained cost functions.} Unlike the former approaches in which the human feedback is implicitly combined in  $\theta_{symbiotic}$ using a human-preferences model, when using direct cost-function adaptation methods the human feedback is explicitly combined in the model's cost function \citep{fsaaai}. Hence, cost-function adaptation can be seen as a flattened version of Equation \eqref{eqn:sym3} where there exist no $\theta_H, \theta_M$ and thus:
\begin{equation}
\hat{\theta}_{symbiotic}= 
\argmin_{\gamma \in \Gamma}
\mathbb{E}_{z\sim d_{human}, d_{data}}[l_{comb}(z;\gamma)],
\end{equation}

which, in its simplest form, translates into:
\begin{equation}
\label{eqn:cost}
l_{comb_i}(x_i, y_i; \gamma) = \argmin_{\gamma \in \Gamma}~ l_1(y_i,h(x_i,\gamma)) + \lambda \times l_2(f_1(y^h_i),f_2(h(x_i,\gamma))).
\end{equation}
 
In Equation \eqref{eqn:cost}, the constraints $\mathcal{C}(\gamma,\hat{\theta}_M) \leq c_1$, $\mathcal{C}(\gamma,\hat{\theta}_H) \leq c_2$ introduced earlier have
have translated into a single hyperparameter, $\lambda$, that explicitly controls the influence of human preferences on the overall learning process. 

A simple application of Equation \eqref{eqn:cost} is when
 $l_1=l_2$, and $f_1=f_2=\mathcal{I}$ where $\mathcal{I}$ represents the exact form of the loss function. In such a case, data points
$y^h_i$ denote human-preference-based labels, which are used to ensure that the predictions align with those preferences. Such an approach will work well if, for instance, there exists multiple, equally-performing decision boundaries, and we choose the one that also highly aligns with human preferences.  However, by parametrizing $l_{comb_i}$ using both $l_1$ and $l_2$, as well as both $f_1$ and $f_2$, the loss can express a richer set of relations between $d_{data}$ and $d_{human}$. An example is when human preference input is not provided on the actual records' labels, but rather on indirect components of the prediction process such as recommended features. In such a case, $f_2$ denotes the important features extracted from a learned model $h_\gamma$, $l_1$ denotes a standard loss function such as cross entropy while $l_2$ denotes a distance metric between the two sets of features.

\section{Recent Advancements in LLMs and Their Cognitive Abilities: Evidence on the Power of Centaurs}\label{sec:llm}

In recent years we have seen a fast growth in the development and use of Generative AI. Notable examples include efforts by OpenIA with GPT (Generative Pre-trained Transformer) and DALL-E (“Dali” and “Eve,” from artist Salvador Dali and the character Eve from the Pixar movie WALL-E), Meta with Llama (Large Language Model Meta AI), Google with Bard that uses LaMDA (Language Model for Dialog Applications) and more recently PaLM (Pathways Language Model), Microsoft with Bing Chat, Stability AI with Stable Diffusion, Github (and OpenAI) with GitHub Copilot, and Anthropic with Claude. 

In 2022, for example, LLMs saw a major advancement: ChatGPT—a large language model
chatbot developed by OpenAI based on GPT-3.5 with the ability to provide conversational responses that
can appear surprisingly human. Like other language models, the main idea behind ChatGPT is simple: to
predict the next word in a sentence or phrase based on the context of previous words, using a model
trained on a large number of instances. GPT-3 had about 175 billion machine learning parameters, and
some estimates showed that it consumed about 936 Mwh to train \citep{energy}---equivalent of about 30k American
households power usage in a day. While recent improvements have focused on making GPT-3 more
efficient by reducing these numbers, other works focused on further improving its reasoning abilities. 
For instance, the “chain of thought prompting” method, first presented by Google Brain \citep{google}, enabled language models of sufficient scale (e.g., models
with 100 billion parameters) to solve complex reasoning problems that are not solvable with
standard prompting methods. In 2023, OpenAI introduced GPT-4, which is not only multimodal (e.g.,
accepts images), but according to the creator “exhibits human-level performance on various professional
and academic benchmarks” \citep{openai}.

Though LLMs have demonstrated extraordinary performance, the exact mechanism behind those impressive achievements is not entirely clear and has been the subject of an ongoing debate. Another, topic sparking ongoing debate is whether, and under which terms, LLMs can be used as \textit{cognitive models}. A cognitive model is defined as ``an approximation of one or more cognitive processes in humans or other animals for the purposes of comprehension and prediction'' \citep{cogdef}. Thus, a model's cognitive abilities can be captured either internally, by the extent to which the model's internal learning and inference processes---captured via various neural activities---resemble those of humans; or externally, by the extent to which the model's output follows human behaviors, decisions, and preferences.  We claim that (a) modern LLMs already possess traits that highly resemble important human cognitive processes, and (b) a main reason for this is that they have moved from pure AI models to centaurs by benefiting from direct incorporation of human intuition as well as symbiotic learning---the two main characteristics of centaurs.  
Building on these traits and enhancing them via more advanced symbiotic learning approaches or further incorporation of human intuition 
 is an important building block for future endeavors in 
 turning LLMs into much stronger cognitive models, capable of further representing human behavior.

The first cognitive-like mechanism underlying LLMs is \textit{attention}. Attention is a fundamental mechanism of cognition needed to perform complex reasoning tasks. Modern cognitive research defines attention as the capacity to select and enhance certain aspects of currently processed information while suppressing the remaining aspects \citep{visual}.
%It can be thought of as a solution to a fundamental computational trade-off that limited agents face in complex environments: on the one hand, the need to focus on as much information as possible, and on the other hand, the necessity to optimize performance due to limited cognitive resources by focusing on the most relevant events \cite{visual}.  
%The mechanism of cognitive attention is not yet fully understood and there exist multiple theories aiming at explaining it. Regardless of the specific theory one chooses to adopt, there exists a consensus among theorists on the existence of two types of attention: \textit{bottom-up} attention and \textit{top-down} attention, and on the fact that the focus of attention is determined by an interaction between the two attention types (though the exact interaction is not yet fully understood) \cite{topbottom,topbottom2,topbottom3}.
%Bottom-up attention refers to the involuntary and unexpected allocation of attention to objects in the environment such as noticeable external stimuli because of their inherent properties relative to other objects in the environment.
%Top-down attention refers to the voluntary allocation of attention to certain objects. Unlike bottom-up attention, it is not stimulus-inspired, but is rather an internally induced process based on prior knowledge or well-defined goals. 
When one attends to an object, she implicitly assigns a higher importance score to that object compared to other objects in the environment.
%; thus, attention can be seen as a dynamic ``filtering'' mechanism which removes unwanted or irrelevant information by assigning irrelevant objects lower importance scores compared to other objects in the environment. 
The higher-scored objects receive preferential processing, leading to an increased neural response which results in improved functionalities on the attended object such as better memory storage \citep{selectiveatt}.
%Another consensus among cognition theorists is on the effects of cognitive attention: when one attends an object, she implicitly assigns a higher importance score to that object compared to other objects in the environment; thus, attention can be seen as a dynamic ``filtering'' mechanism which removes unwanted or irrelevant information by assigning irrelevant objects lower importance scores compared to other objects in the environment. 
%The higher-scored objects receive preferential processing, leading to an increased neural response which results in  improved functionalities on the attended object such as better memory storage \cite{selectiveatt}.
%and thus optimizes a human's action to achieve the current goal (Desimone and Duncan, 1995). It is crucial as well for intelligent agents to integrate and utilize external and internal information efficiently and to reach a signal-to-noise ratio

Relative importance-score assignment is also the core objective of the attention mechanism used as the key novelty in the Transformer architecture---the neural-network (NN) architecture underlying the vast majority of modern LLMs \citep{transformer}. %In a nutshell, the attention mechanism aims at learning the \textit{relative} importance of a set of tokens $\{x_j\}$ in one sequence (which can be either the input text, output text, or both) to a given token, $x$, in the same/another sequence. 
The core idea of artificial attention---a set of comparisons to relevant items in some context, a normalization of those scores to provide a probability distribution, followed by a weighted contextualization based on those scores---closely resembles the human cognitive attention process, in which humans assign importance scores to objects in the environment, where those scores are both relative and dynamically learned by the human.  

Furthermore, the usage of attention for creating contextualized representations of objects based on prior relevant ones resembles another cognitive process, \textit{priming}, occurring when the exposure to one stimulus influences a response to a subsequent stimulus. Specifically, the Transformer's use of attention weights for the creation of contextualized representations resembles a specific type of priming, Contextual Priming \citep{contprim} in which humans use a context to speed up processing for stimuli that are likely to occur in that context.
Indeed, studies in human language confirm that while reading, people will look back at previous sections of text in order to clarify what they are currently reading \citep{primingevidence,reading2}.

While the attention mechanism as used in modern LLM architectures resembles human cognitive processes in many respects, artificial attention mechanisms as described above and cognitive attention vary on a number of dimensions. For instance, while cognitive attention is manifested via both top-down and bottom-up attention mechanisms, artificial attention mechanisms such as the one used in the Transformer architecture are inherently goal-driven and are thus solely top-down based. Moreover, the attention model as currently used in most NN architectures computes an attention score for a given query for every possible key, whereas cognitive attention is ``dynamically selective:'' the human's choice of which objects to attend depends on the human's current state \citep{principlepsych}.  
One possible way to bridge those gaps between artificial attention and human attention is a \textit{symbiotic learning-based attention mechanism}. In such a symbiotic learning process, human attention preferences as extracted from real-world observational data on tasks such as reading and text comprehension are incorporated in the learned attention matrices 
%$W_s$ 
as either hard or soft constraints, creating a centaur Transformer model. Such human attention preferences can both incorporate bottom-up insights that current attention mechanisms in the NLP domain are not able to explicitly express due to the intrinsic guided learning process, as well as provide attention insights that dynamically depend on the human's current state. 
%We note that, while such symbiotic learning approach will probably lead to an artificial attention mechanism that is more similar to the cognitive attention process, it is not immediately clear whether it will also lead to more accurate models. We elaborate on this idea in Section \ref{sec:discussion}.

%While works aiming at understanding human attention comaed to artifi attention in the nlp domain is scarce,  , some very recent works in the vision domain have tried to quantify empirical differences between the twp using crowdsourcing platforms such mturk. We elaborate on the concuion and finding of those works in section ?. 

\vspace{5pt}
When GPT-3 was first launched, its creators noticed a rather interesting behavior: the pre-trained model is able to adapt itself to a variety of downstream tasks without any further training---\textit{i.e.}, without changing the model's weights using task-specific data \citep{fewshot}. Instead, all that was needed in order to obtain a high degree of accuracy on the downstream task is a well-designed prompt---a natural language description of the task---coupled with a very small number of demonstrations of object-label examples (``few shot learning''). %If the number of demonstrations is zero and only a prompt is available, we refer to the interaction as ``zero-shot learning''. This  unexpected observation of behavior adaptation without weight adaptation is rather surprising, and yet to be fully understood. Multiple works have tried to understand the in-context learning mechanism. 
%[] claims that llm's strong in-context learning abilities emerge as a by-product of enforcing the network to represent the input sequential data through its bounded memory architecture [Brown et al. 2020]. 
%The leading paradigm for explaining the high performance of in-context learning in LLMs is a Bayesian paradigm \cite{latent,latent2}, interpreting in-context learning as the process of locating latent concepts the LLM has already acquired from pretraining data. The LLM uses the prompt provided during in-context learning to locate a previously learned concept where a ``concept'' is either a single latent variable or a combination of many latent variables that specify different aspects of the semantics and syntax of a document.
%The process of locating learned concepts is formulated as bayesian inference of a prompt concept that is mutual to the prompt's examples, as seen in Equation \ref{eqn:bayes}. If the model can infer the prompt's main concept, then it can be used to make correct output predictions on the test example. 
%\begin{equation} 
%\label{eqn:bayes}
%p(output|prompt) = \int_{concept} p(outout|prompt,concept)p(concept|prompt)p(concept)
%\end{equation}
Interestingly, the idea of in-context learning in LLMs strongly reassembles the idea of Concept Learning, a critical aspect of cognition. Concept learning was first defined by \citep{conceptfirst} as ``the search for and listing of attributes that can be used to distinguish exemplars from non exemplars of various categories.'' Concept learning resembles LLM's in-context learning in two key points: the ability to obtain abstract concepts from examples, and the ability to learn from very few examples. 
As shown in prior work, humans are able to acquire the definition of a general category from
given sample positive and negative training examples of the category \citep{baystwo,another}.
As further shown in prior work, humans can learn such general-category definitions using only a few examples \citep{small,small2}. %In specific domains such as the vision domain, it has been shown that humans are able to acquire visual concepts without \textit{any} visual examples, by relying on cross-modal transfer from language descriptions to the visual domain \cite{smallzero,smallzero2}.

The learning process in the classic in-context-learning setting is a one-way learning process in which the human ``teaches'' the machine by designing prompts and demonstrations, fed to the LLM which outputs the required text according to the human-designed prompt. While such one-way in-context learning schemes were shown to yield high performance, recent work has shown that in-context learning does not result in true model learning, but is rather merely a sophisticated ``query normalization'' method for building better encoding of user queries \citep{prompt1,prompt2}. An interesting question that should be explored is whether symbiotic learning can be used to improve existing in-context learning approaches, facilitating true learning by enhancing existing one-way in-context learning schemes into two-way schemes. Two-way in-context learning schemes result in centaur models in which both the human and the machine learn from each other's inputs, instead of only the machine learning from the human. A concrete example of such a two-way learning process for in-context learning is symbiotic prompt learning as illustrated below: instead of the human manually crafting prompts and examples, a human can use a machine-learning model, $h_1$, that, given a general task description and goals inputted by the human, $d_{human}$,  outputs a prompt and a set of demonstrations, $d_{prompt}$ designed to be consumed by a specific type of LLM, $h_2$:

\vspace{5pt}
Human ($\rightarrow d_{human}$) $\rightarrow$ Machine ($\rightarrow d_{prompt}=h_1(d_{human})$) $\rightarrow$ Machine ($\rightarrow h_2(h_1(d_{human}))$).

\vspace{5pt}
The interactions between the human and the machine can become even more flexible by enabling the human to give feedback ($\mathcal{H}$) on the prompts and demonstrations outputted by the machine before inputting them to the LLM:

\vspace{5pt}
Human ($\rightarrow d_{human}$) $\rightarrow$ Machine ($\rightarrow d_{prompt}=h_1(d_{human})$) 
$\rightarrow$
Human ($\rightarrow \mathcal{H}(h_1(d_{human}))$)
$\rightarrow$ Machine ($\rightarrow h_2( \mathcal{H}(h_1(d_{human})))$).

\vspace{10pt}
{When faced with complex problems, one reasoning strategy employed by humans' cognition is ``divide-and-concur:" decomposing the problem into manageable, intermediate steps and solving each one
before giving the final answer \citep{wei2022chain}. Such intermediate sub-problems can be seen as a series of logical deductions that sequentially lead to a conclusive answer.
The benefits of such sequential decomposition reasoning in humans raised the question of whether such ``intermediate thinking" patterns can improve LLMs' reasoning abilities, leading to the development of Chain-of-Thought reasoning \citep{wei2022chain}. In its simplest form, Chain-of-Thought (CoT) denotes
a series of intermediate natural language reasoning steps that lead to a final output \citep{reasoning}.}

{Until recently, CoT has been incorporated within LLMs primarily using 
CoT Prompting: instead of demonstrations solely composed of a (query, answer) pair, CoT Prompting additionally includes a rationale for each example, encouraging the model to verbalize the intermediate reasoning steps for solving the task \citep{reasoning}.
Newer models, such as the latest release of OpenAI’s o1 \citep{o12}, take one step forward and incorporate CoTs within the model's learning and reasoning process. In this way, human input in the form of CoTs is explicitly combined  within the model's architecture rather than being queried in an ad-hoc fashion by the end user as part of the prompt. The o1 model is trained to follow human-produced CoTs by using an RL scheme designed to improve the model's alignment not only on final answers but also on their corresponding CoTs---the reasoning process that led to the final answers \citep{planning,o11}.}  

%{While CoT prompting/learning has been empirically proven to significantly improve LLMs' reasoning abilities \citep{cot1}, it has some fundamental limitations. For instance, it has been suggested \citep{unfaithful1,unfaithful2} that CoT explanations are \textit{unfaithful} --- they neither accurately represent the specific reasons behind model predictions nor the model's high level reasoning process. The latter poses the risk of falsely trusting AI systems due to the alleged plausibility of explanations --- which, as plausible as they may be, may not accurately represent the true reasoning process of the model.}

%\textcolor{blue}{Perhaps the biggest limitation of training LLMs on CoT explanations is that the they are written by humans. As such, they can be misleading, biased, omitting crucial parts of the causal chain for a particular event, or be unfaithful accounts of the human's own cognitive process \citep{unfaithful1,cotnogood1,cotnogood2}. Moreover, such human-produced CoTs can, by definition, only represent \textit{conscious} reasoning patterns, depriving LLMs from any training data representing \textit{unconscious reasoning patterns.} This is in contrast to observed signals of human intuition, such as actions or preferences, of which both conscious and unconscious reasoning patterns can be extracted.}

While CoT reasoning can certainly be seen as one step forward towards creating centaur models, as mentioned earlier, such models are still different than centaurs. A critical difference is in that they include only \textit{conscious} reasoning patterns, depriving LLMs from any training data representing \textit{unconscious reasoning patterns.} Thus, models trained on CoT human input alone may not be considered as centaurs: they lack true human intuition, at least based on the definition of intuition we discussed earlier.

Indeed, recent works show that while o1 significantly outperforms GPT-4 on problem-solving tasks such as math and coding problems, it performs on par (and sometimes, subpar) with GPT-4 on ambiguous tasks, highly-abstract tasks, tasks requiring flexibility an adaptability to real-time dynamics, or ones that require knowledge gained through experience \citep{o11,o12}. Furthermore, the iterative reasoning process of o1 makes it significantly slower than GPT-4, leading to longer output times \citep{o11}.

Nontheless, enriching LLMs' built-in, cognitive-like mechanisms such as attention and concept learning with symbiotic learning techniques and high-quality human-preference data has a great potential to further enrich LLMs' cognitive abilities. This can be operationalized, for example, by using two-way in-context learning and human-preference-based attention weights as described above. 
Below we discuss two specific symbiotic-learning methods that have shown to be the underlying reasons why LLMs have gained
cognitive abilities: human-preference-based supervised fine-tuning and human-guided rewards.

Fine-tuning LLMs with human-preference data has been shown by multiple works to yield models that successfully follow human behaviors on a diverse array of applications. All works apply a similar technique to the one illustrated in Section \ref{sec:symbiotic}, while varying the nature of $d_{data}$ and $d_{human}$ as well as the structure of $\gamma$ and $\mathcal{C}$ to meet the needs of the specific downstream task and using efficient fine-tuning techniques such as linear probing or LORA adaptation.  For example, \citep{reading,reading2} showed that LLMs can predict behavioral responses in reading tasks by fine-tuning pre-trained transformers on both behavioral and neural human reading data, fitting a regression model from the model activations' to the corresponding neural/behavioral measurements. Furthermore, \citep{motion} showed that LLMs can follow human motion behavior by fine-tuning a pre-trained LLM on human motion data using LORA adaptation \citep{lora}, and \citep{binz} discussed that LLMs can imitate human's decision-making process by fine-tuning pre-trained LLMs on human-choice datasets, fitting a regression model from the model's activations to the corresponding human choices. By doing so, \citep{binz} created a new class of models that they termed CENTaUR. The resulting model showed close to human behavior in two general sets of human decision-making experiments. The first set is known as decisions from descriptions. In this set, a complete, idealized, and abstract set of information about the values and frequencies of potential outcomes from each choice is provided to participants before choices are made \citep{26}. An example is when participants are told there is “50\% chance to win 1000\$; 50\% chance to win nothing” (\citep{27}, p. 264). The second set is known as decisions from experience. In this set, participants need to form their own view of the potential outcomes from each choice via feedback provided after each selection is made \citep{26,binz}.

%cognitive models-also neiral activity
Human-guided rewards is another symbiotic learning approach that has been used to enhance
 LLMs' cognitive abilities. Human-preference-based reward schemes were shown by multiple works \citep{firstllmreward1,firstllmreward2} as crucial for unsupervisedly pre-trained LLMs to create outputs that better align with human's choices. Newer works have also applied human-guided reward schemes to increase LLMs' alignment with collaborative human decision-making mechanisms \citep{collaborative}, to create LLMs that both follow and predict humans' implicit and explicit intentions \citep{intentions}, or to teach LLMs how to align with humans' moral values, such as honesty and fairness \citep{values}.

Other works such as \citep{binz2} and \citep{28} show that LLMs have
been able to perform well compared
to human subjects in vignette-based tasks, and has also excelled in multi-arm bandit type of decision making experiments that requires balancing exploitative and exploratory actions. Some specific
cognitive experiments performed by researchers to assess the cognitive ability of LLMs include:

\textbf{Base Rate Fallacy (The Cab Problem).} This experiment is about judging the color of a cab that has been involved in an accident. Base rate fallacy refers to the fact that participants often fail to take into account the base rate of different cab colors that operate in a city.  GPT has been able to avoid this fallacy, providing accurate answers \citep{binz2}.

\textbf{Causal and Counterfactual Reasoning (Blicket, Pills, and Do Experiments).} One of the experiments used in assessing the causal reasoning ability of GPT is the “Blicket” experiment. The goal is to identify whether an object turns on a machine (\textit{i.e.}, is a “blicket”). Two objects are introduced, and the participants are informed that the first object alone can turn on the machine. But for the second object to turn on the machine, it needs to be accommodated with the first one. GPT-3 has been able to correctly identify that the first object is a blicket, but the second one is not \citep{binz2}. A more serious set of investigations includes using the so-called “pills experiments” and “do experiments”. In both of these, an underlying causal relationship between different entities (e.g., pills and death) is provided and then the participant is asked some counterfactual questions. GPT-3 without fine-tuning has shown good performance in some of these experiments, but has also raised concerns about its overall causal and counterfactual reasoning. For example, authors in \citep{binz2} concluded that “GPT-3 has difficulties with causal reasoning in tasks that go beyond a vignette-based characterization.” Using experiments related to over hundred causal relationships from various domains such as physics, biology, zoology, cognitive science, epidemiology, and soil science, authors in \citep{28} concluded that “algorithms based on GPT-3.5 and 4 outperform existing algorithms on a pairwise causal discovery task, counterfactual reasoning task, and actual causality.  At the same time, LLMs exhibit unpredictable failure modes.”

\textbf{Problem Solving and Decision-Making (Multi-Arm Bandits and Prospect Theory Experiments).} GPT has also shown strong performance in problem-solving and decision-making tasks \citep{binz2}. As mentioned earlier, these include both decisions from descriptions and decisions from experience. The latter is the more challenging one, and often requires some fine-tuning to make sure human intuition is directly fed to the underlying model. In testing the ability of LLMs to make decisions from experience, some researchers have focused on multi-arm bandit problems where tradeoffs in exploration and exploitation play a key role. Recent work also demonstrates that LLMs might show similar to human behavior biases such as those that were documented by Kahneman and Tversky as part of their celebrated Prospect Theory \citep{27,36}. For example, in one set of experiments researchers found that GPT-3 showed three of the six biases in human decision-making that Prospect Theory predicts. Specifically, GPT-3 displayed a “framing effect”, a “certainty effect,” and an “overweighting bias effect” but not a “reflection effect,” an “isolation effect,” or a “magnitude perception effect” \citep{binz2}.

Finally, a particularly interesting work is that of \citep{gpt4better}, which aims at evaluating the cognitive abilities of different LLMs by probing LLMs with the Cognitive Reflection Test and comparing their performance and mistakes. The authors found that older versions of GPT made significantly more cognitive errors compared to GPT-4. The authors discuss potential reasons for such cognitive differences and conclude that the heavier reliance on human-guided rewards (using RLHF) for model training is the most plausible reason for GPT-4's improved cognitive abilities. 

As discussed earlier, these methods are all based on symbiotic learning. Furthermore, they allow the incorporation of human intuition so as to augment performance in various tasks. As such, LLMs that benefit from these techniques have quietly moved to become centaurs. It is not clear whether and how without becoming centaurs, LLMs could reach such cognitive ability.

\section{Discussion: Are Centaurs Always Superior?
}\label{sec:discussion}

In prior sections we have surveyed multiple techniques for creating centaurs.
%by incorporating 
%human intuition and preferences into AI models, as well as proposed future directions in the research of such techniques.
A key question that arises when considering the deployment of centaurs is the following:
Does the fact that centaurs incorporate human intuition and symbiotic learning make them always have a better performance vis-a-vis some pure AI models?

The answer to this question is simple but important and requires a holistic view of the specific application as well as the performance metrics used. 
In some applications, there is a strong positive correlation between the model's alignment with human preferences and high performance as dictated by the task-specific performance metrics. In some other applications,
such a correlation is weak at best, or even negative. Thus, we claim that there exists an inherent tension between the model's alignment with human preferences and the model's performance, and represent this tension by evaluating the model using two separate evaluation metrics: a performance metric, $\phi_p$, and a behavioral metric, $\phi_b$. We see centaurs as a key conceptual mechanism over the next few years to create AI models that perform well in both $\phi_p$ and $\phi_b$ dimensions.

However, ensuring that future centaurs perform well in both dimensions needs further research and deeper understandings. To illustrate this, we consider current efforts for increasing LLMs' cognitive abilities as discussed in Section \ref{sec:llm}.
%Creating stimulus-computable cognitive models that are able to follow both human neural activity and human behavioral activity has been one of the most desirable goals of AI research ever since its emergence in the 1950s. Human cognition, however, has its limitations: memory constraints, difficulty in processing large amounts of complex information, and built-in biases \cite{cogdef}. 
Turning LLMs into cognitive models means that LLMs will also inherit at least some flaws of human cognition, not just its strengths.
Indeed, it has been shown that on some tasks, LLMs' ability to follow human's intuitions,  behaviors, or decisions negatively correlates with the model's performance \citep{studies2,bias1}. That is to be expected as human decision-making process is oftentimes irrational, influenced by noise factors such as emotions and cognitive biases which might result in suboptimal decisions and behaviors \citep{emotion}. For instance, it has been shown in \citep{bias1}, \citep{bias2}, and \citep{bias3} that LLMs share similar cognitive biases with humans, and that even though those biases lead to behaviors that are more similar to human behaviors, the biased behaviors also lead to deterioration in LLMs' performance. While the latter has been recently highlighted in the context of LLMs, the cognitive bias problem presents a challenge in the design of any AI model where human biases are combined in the training process either explicitly by combining human preferences in the training algorithm, or implicitly by using biased training sets \citep{fairness}. {Another potential threat in developing centaur-type models driven by human intuition is the development of dark-triad-like traits \citep{darktriad} within LLMs. At scale, such traits may pose serious safety risks by driving the model to engage is an array of negative behaviors from systematic deception and content manipulation to power-seeking behaviors, potentially with the collaboration of other centaur-like agents \citep{risk1,risk2,risk3}.}

Nonetheless, there exist multiple settings in which we can expect centaurs to score well on both $\phi_p$ and $\phi_b$ dimensions. This is clearly the case in which these dimensions are strongly and positively correlated, such as developing LLM-based chatbots \citep{intentions} in which the performance metric is the extent to which human customers find them ``helpful.'' Such a performance metric is heavily human-aligned, and thus different from an ``objective'' performance metric such as the proportion of chatbot's answers that contained factually correct information. Likewise, the performance metric of many summarizing LLMs \citep{firstllmreward1,firstllmreward2} is whether the produced summary is highly-rated by human readers---a purely behavioral metric, as opposed to a factual metric such as the number of grammar mistakes found in the summary. 

Another type of tasks in which we can expect centaurs to score well on both $\phi_p$ and $\phi_b$ is tasks in which the model's goal is merely to accurately replicate human behavior. For example, multiple works have examined the extent to which LLMs can be used as a tool for simulating human behaviors in human subject studies \citep{studies,studies2,studies3}. It has been shown that certain LLMs can accurately reproduce economic, psycholinguistic, and social psychology studies, and existing research efforts are targeted at generalizing those results by using LLMs for simulating  diverse populations, both in terms of their simulated demographic characteristics and their simulated personality traits \citep{personality1,personality2}.  

Other notable examples include using centaurs in tasks for
 which there exist multiple optimal solutions with respect to $\phi_p$. Benefiting from human intuition and symbiotic learning, a centaur model can pick
 an optimal solution that best aligns with human preferences. A similar situation exists in 
 tasks applied to unrepresentative datasets or datasets with many edge cases.

We consider modern centaur-based symbiotic learning to be a key mechanism for ensuring that LLMs in particular and AI models in general score high on both the performance metric and the behavioral metric for a variety of reasons. A particular reason is that by carefully choosing $\gamma$ and $\mathcal{C}$, the model is able to learn when following human intuition is beneficial, while selectively filtering damaging intuitions resulting from irrational patterns such as cognitive biases and emotions. Alternatively, if the performance metric is merely a behavioral metric, as is the case when using LLMs for behavior simulation in human subject studies, $\gamma$ and $\mathcal{C}$ can be fine-tuned to solely capture human intuition. Evidence of our hypothesis can already be found in existing work,
and as the research of centaurs continues, more sophisticated ways for balancing performance-based evaluation metrics and behavior-based evaluation metrics will emerge, creating ``selectively cognitive'' models that are able to learn when and how it is best to follow human intuition.

\section{Conclusion: 
The Future of AI In Various Applications is Centaur 
}\label{sec:conclusion}

In this article, we presented the concept of centaurs as AI models that benefit from the paradigm of symbiotic learning as well as the incorporation of human intuition. We argued that centaurs have various advantages over other AI models. We also discussed that various human-level performances observed from recent LLMs including demonstration of cognitive ability stem from the fact that they have moved from pure AI models to centaur-based systems.

We started by presenting a wide array of motivations for the incorporation of human intuition in training AI models. We then proceeded to surveying traditional human-in-the-loop learning techniques and explained how they differ from symbiotic learning techniques which are used in centaurs. We then presented a formal framework for symbiotic learning and discussed five concrete symbiotic learning techniques: preference-based augmented covariate space, human-guided rewards, human-preference-based supervised fine-tuning, human-machine ensembles, and preference-constrained cost functions. 
%To illustrate the importance of symbiotic learning, we showed how various symbiotic learning techniques can enhance LLMs' baseline cognitive abilities, resulting from cognitive-like learning processes such as attention and in-context learning. We concluded with a discussion of the limitations of human intuition incorporation in ML models' training, and articulated how symbiotic learning may serve as a key paradigm for overcoming those limitations. 

In closing, we note that while AI is still far from reaching “human-level intelligence,” the advancements in Generative AI have shown promising results for making use of centaurs in various applications. Centaurs might indeed be at the center of future AI developments. They might also be our main hope to get closer to reaching the goals that were stated by the founding fathers of AI. To realize this, it is useful to revisit what the founding fathers (John McCarthy, Marvin Minsky, Nathaniel Rochester, and Claude Shannon) stated in 1955 as part of their proposed 2-month, 10-man study of AI (to be held at Dartmouth):

``The study is to proceed on the basis of the conjecture that every aspect of learning or any other feature of intelligence can in principle be so precisely described that a machine can be made to simulate it. An attempt will be made to find how to make machines use language, form abstractions and concepts, solve kinds of problems now reserved for humans, and improve themselves'' \citep{34}.

Making ``machines use language, form abstractions and concepts, solve kinds of problems now reserved for humans, and improve themselves'' no longer seems an unrealistic goal, given what modern centaur-based AI methods have been able to achieve to this date. Future research can make these methods even more powerful, pushing us further to reach the goals of AI research as envisioned by the founding fathers of AI.
\printbibliography

@article{rewardensemble,
  title={Improving Reinforcement Learning from Human Feedback with
Efficient Reward Model Ensemble},
  author={Zhang and Zhenfang Chen and Sunli Chen and Yikang Shen and Zhiqing Sun and Chuang Gan},
  journal={arXiv preprint arXiv:2401.16635 },
  year={2024}
}

@inproceedings{fsaaai,
  title={Human-in-the-loop feature selection},
  author={Correia, Alvaro HC and Lecue, Freddy},
  booktitle={Proceedings of the AAAI Conference on Artificial Intelligence},
  volume={33},
  number={01},
  pages={2438--2445},
  year={2019}
}

@article{prior,
  title={Human-in-the-loop interpretability prior},
  author={Lage, Isaac and Ross, Andrew and Gershman, Samuel J and Kim, Been and Doshi-Velez, Finale},
  journal={Advances in neural information processing systems},
  volume={31},
  year={2018}
}

@article{teachingf2,
  title={Machine teaching: A new paradigm for building machine learning systems},
  author={Simard, Patrice Y and Amershi, Saleema and Chickering, David M and Pelton, Alicia Edelman and Ghorashi, Soroush and Meek, Christopher and Ramos, Gonzalo and Suh, Jina and Verwey, Johan and Wang, Mo and others},
  journal={arXiv preprint arXiv:1707.06742},
  year={2017}
}

@article{hic2,
  title={How AI-Human symbiotes may reinvent innovation and what the new centaurs will mean for cities},
  author={Muller, Emmanuel},
  journal={Technology and Investment},
  volume={13},
  number={1},
  pages={1--19},
  year={2022},
  publisher={Scientific Research Publishing}
}

@article{hic,
  title={Human-agent collectives},
  author={Jennings, Nicholas R and Moreau, Luc and Nicholson, David and Ramchurn, Sarvapali and Roberts, Stephen and Rodden, Tom and Rogers, Alex},
  journal={Communications of the ACM},
  volume={57},
  number={12},
  pages={80--88},
  year={2014},
  publisher={ACM New York, NY, USA}
}

@article{teachingf1,
  title={On the complexity of teaching},
  author={Goldman, Sally A and Kearns, Michael J},
  journal={Journal of Computer and System Sciences},
  volume={50},
  number={1},
  pages={20--31},
  year={1995},
  publisher={Elsevier}
}

@inproceedings{secteach1,
  title={Using machine teaching to identify optimal training-set attacks on machine learners},
  author={Mei, Shike and Zhu, Xiaojin},
  booktitle={Proceedings of the aaai conference on artificial intelligence},
  volume={29},
  number={1},
  year={2015}
}

@inproceedings{secteach2,
  title={The security of latent dirichlet allocation},
  author={Mei, Shike and Zhu, Xiaojin},
  booktitle={Artificial Intelligence and Statistics},
  pages={681--689},
  year={2015},
  organization={PMLR}
}

@article{eduteach2,
  title={Optimal teaching for limited-capacity human learners},
  author={Patil, Kaustubh R and Zhu, Jerry and Kope{\'c}, {\L}ukasz and Love, Bradley C},
  journal={Advances in neural information processing systems},
  volume={27},
  year={2014}
}

@inproceedings{eduteach1,
  title={Machine teaching: An inverse problem to machine learning and an approach toward optimal education},
  author={Zhu, Xiaojin},
  booktitle={Proceedings of the AAAI conference on artificial intelligence},
  volume={29},
  number={1},
  year={2015}
}

@article{rlhforiginal,
  title={Deep reinforcement learning from human preferences},
  author={Christiano, Paul F and Leike, Jan and Brown, Tom and Martic, Miljan and Legg, Shane and Amodei, Dario},
  journal={Advances in neural information processing systems},
  volume={30},
  year={2017}
}

@article{cl1,
  title={Competence-based curriculum learning for neural machine translation},
  author={Platanios, Emmanouil Antonios and Stretcu, Otilia and Neubig, Graham and Poczos, Barnabas and Mitchell, Tom M},
  journal={arXiv preprint arXiv:1903.09848},
  year={2019}
}

@article{cl2,
  title={Learning the curriculum with bayesian optimization for task-specific word representation learning},
  author={Tsvetkov, Yulia and Faruqui, Manaal and Ling, Wang and MacWhinney, Brian and Dyer, Chris},
  journal={arXiv preprint arXiv:1605.03852},
  year={2016}
}

@article{cl3,
  title={Curriculum learning and minibatch bucketing in neural machine translation},
  author={Kocmi, Tom and Bojar, Ondrej},
  journal={arXiv preprint arXiv:1707.09533},
  year={2017}
}

@inproceedings{cl5,
  title={Curriculumnet: Weakly supervised learning from large-scale web images},
  author={Guo, Sheng and Huang, Weilin and Zhang, Haozhi and Zhuang, Chenfan and Dong, Dengke and Scott, Matthew R and Huang, Dinglong},
  booktitle={Proceedings of the European conference on computer vision (ECCV)},
  pages={135--150},
  year={2018}
}

@article{humaneval1,
  title={Reading tea leaves: How humans interpret topic models},
  author={Chang, Jonathan and Gerrish, Sean and Wang, Chong and Boyd-Graber, Jordan and Blei, David},
  journal={Advances in neural information processing systems},
  volume={22},
  year={2009}
}

@article{cl4,
  title={Stc: A simple to complex framework for weakly-supervised semantic segmentation},
  author={Wei, Yunchao and Liang, Xiaodan and Chen, Yunpeng and Shen, Xiaohui and Cheng, Ming-Ming and Feng, Jiashi and Zhao, Yao and Yan, Shuicheng},
  journal={IEEE transactions on pattern analysis and machine intelligence},
  volume={39},
  number={11},
  pages={2314--2320},
  year={2016},
  publisher={IEEE}
}

@article{humaneval2,
  title={Deep generative image models using a laplacian pyramid of adversarial networks},
  author={Denton, Emily L and Chintala, Soumith and Fergus, Rob and others},
  journal={Advances in neural information processing systems},
  volume={28},
  year={2015}
}

@inproceedings{anomalyensemble,
  title={Towards Unsupervised Validation of Anomaly Detection Models},
  author={Lihi Idan},
  booktitle={27th European Conference on Artififcial Intelligence (ECAI)},
  year={2024},
doi          = {10.3233/FAIA240859}
}

@article{calibration,
  title={Human-aligned calibration for ai-assisted decision making},
  author={Corvelo Benz, Nina and Rodriguez, Manuel},
  journal={Advances in Neural Information Processing Systems},
  volume={36},
  year={2024}
}

@inproceedings{cloriginal,
  title={Curriculum learning},
  author={Bengio, Yoshua and Louradour, J{\'e}r{\^o}me and Collobert, Ronan and Weston, Jason},
  booktitle={Proceedings of the 26th annual international conference on machine learning},
  pages={41--48},
  year={2009}
}

@article{lora,
  title={Lora: Low-rank adaptation of large language models},
  author={Hu, Edward J and Shen, Yelong and Wallis, Phillip and Allen-Zhu, Zeyuan and Li, Yuanzhi and Wang, Shean and Wang, Lu and Chen, Weizhu},
  journal={arXiv preprint arXiv:2106.09685},
  year={2021}
}

@inproceedings{adapter1,
  title={Parameter-efficient transfer learning for NLP},
  author={Houlsby, Neil and Giurgiu, Andrei and Jastrzebski, Stanislaw and Morrone, Bruna and De Laroussilhe, Quentin and Gesmundo, Andrea and Attariyan, Mona and Gelly, Sylvain},
  booktitle={International conference on machine learning},
  pages={2790--2799},
  year={2019},
  organization={PMLR}
}

@article{adapter2,
  title={Learning multiple visual domains with residual adapters},
  author={Rebuffi, Sylvestre-Alvise and Bilen, Hakan and Vedaldi, Andrea},
  journal={Advances in neural information processing systems},
  volume={30},
  year={2017}
}

@inproceedings{inc2,
  title={Interactive machine learning},
  author={Fails, Jerry Alan and Olsen Jr, Dan R},
  booktitle={Proceedings of the 8th international conference on Intelligent user interfaces},
  pages={39--45},
  year={2003}
}

@article{inc4,
  title={Ilastik: interactive machine learning for (bio) image analysis},
  author={Berg, Stuart and Kutra, Dominik and Kroeger, Thorben and Straehle, Christoph N and Kausler, Bernhard X and Haubold, Carsten and Schiegg, Martin and Ales, Janez and Beier, Thorsten and Rudy, Markus and others},
  journal={Nature methods},
  volume={16},
  number={12},
  pages={1226--1232},
  year={2019},
  publisher={Nature Publishing Group US New York}
}

@article{inc5,
  title={Visual classifier training for text document retrieval},
  author={Heimerl, Florian and Koch, Steffen and Bosch, Harald and Ertl, Thomas},
  journal={IEEE Transactions on Visualization and Computer Graphics},
  volume={18},
  number={12},
  pages={2839--2848},
  year={2012},
  publisher={IEEE}
}

@article{incfirst,
  title={Interactive machine learning: letting users build classifiers},
  author={Ware, Malcolm and Frank, Eibe and Holmes, Geoffrey and Hall, Mark and Witten, Ian H},
  journal={International Journal of Human-Computer Studies},
  volume={55},
  number={3},
  pages={281--292},
  year={2001},
  publisher={Elsevier}
}

@article{inc1,
  title={Power to the people: The role of humans in interactive machine learning},
  author={Amershi, Saleema and Cakmak, Maya and Knox, William Bradley and Kulesza, Todd},
  journal={AI magazine},
  volume={35},
  number={4},
  pages={105--120},
  year={2014}
}

@inproceedings{active1,
  title={From theories to queries: Active learning in practice},
  author={Settles, Burr},
  booktitle={Active learning and experimental design workshop in conjunction with AISTATS 2010},
  pages={1--18},
  year={2011},
  organization={JMLR Workshop and Conference Proceedings}
}

@inproceedings{healthact1,
  title={Deep bayesian active learning with image data},
  author={Gal, Yarin and Islam, Riashat and Ghahramani, Zoubin},
  booktitle={International conference on machine learning},
  pages={1183--1192},
  year={2017},
  organization={PMLR}
}

@book{reading2,
  title={Attention and performance XII: The psychology of reading},
  author={Coltheart, Max},
  year={2016},
  publisher={Routledge}
}

@article{healthact2,
  title={Active learning with support vector machines in the drug discovery process},
  author={Warmuth, Manfred K and Liao, Jun and R{\"a}tsch, Gunnar and Mathieson, Michael and Putta, Santosh and Lemmen, Christian},
  journal={Journal of chemical information and computer sciences},
  volume={43},
  number={2},
  pages={667--673},
  year={2003},
  publisher={ACS Publications}
}

@article{financeact2,
  title={Streaming active learning strategies for real-life credit card fraud detection: assessment and visualization},
  author={Carcillo, Fabrizio and Le Borgne, Yann-A{\"e}l and Caelen, Olivier and Bontempi, Gianluca},
  journal={International Journal of Data Science and Analytics},
  volume={5},
  pages={285--300},
  year={2018},
  publisher={Springer}
}

@inproceedings{financeact1,
  title={Machine learning methods to detect money laundering in the bitcoin blockchain in the presence of label scarcity},
  author={Lorenz, Joana and Silva, Maria In{\^e}s and Apar{\'\i}cio, David and Ascens{\~a}o, Jo{\~a}o Tiago and Bizarro, Pedro},
  booktitle={Proceedings of the first ACM international conference on AI in finance},
  pages={1--8},
  year={2020}
}

@article{work2,
  title={Classification with reject option in gene expression data},
  author={Hanczar, Blaise and Dougherty, Edward R},
  journal={Bioinformatics},
  volume={24},
  number={17},
  pages={1889--1895},
  year={2008},
  publisher={Oxford University Press}
}

@article{work4,
  title={Experimental evidence of effective human--AI collaboration in medical decision-making},
  author={Reverberi, Carlo and Rigon, Tommaso and Solari, Aldo and Hassan, Cesare and Cherubini, Paolo and Cherubini, Andrea},
  journal={Scientific reports},
  volume={12},
  number={1},
  pages={14952},
  year={2022},
  publisher={Nature Publishing Group UK London}
}

@article{work3,
  title={A framework for reasoning about the human in the loop},
  author={Cranor, Lorrie F},
  year={2008},
  publisher={Carnegie Mellon University}
}

@article{work1,
  title={How AI can learn from the law: putting humans in the loop only on appeal},
  author={Cohen, I Glenn and Babic, Boris and Gerke, Sara and Xia, Qiong and Evgeniou, Theodoros and Wertenbroch, Klaus},
  journal={npj Digital Medicine},
  volume={6},
  number={1},
  pages={160},
  year={2023},
  publisher={Nature Publishing Group UK London}
}

@article{binz,
  title={Turning large language models into cognitive models},
  author={Binz, Marcel and Schulz, Eric},
  journal={arXiv preprint arXiv:2306.03917},
  year={2023}
}

@book{principlepsych,
  title={The principles of psychology},
  author={Spencer, Herbert},
  volume={1},
  year={1870},
  publisher={Williams and Norgate London}
}

@incollection{primingevidence,
title = {Attention in Language},
booktitle = {Neurobiology of Attention},
publisher = {Academic Press},
pages = {324-329},
year = {2005},
author = {Andriy Myachykov and Michael I. Posner},
}

@article{visual,
  title={Visual attention.},
  author={Allport, Alan},
  year={1989},
  publisher={The MIT Press}
}

@article{transformer,
  title={Attention is all you need},
  author={Vaswani, Ashish and Shazeer, Noam and Parmar, Niki and Uszkoreit, Jakob and Jones, Llion and Gomez, Aidan N and Kaiser, {\L}ukasz and Polosukhin, Illia},
  journal={Advances in neural information processing systems},
  volume={30},
  year={2017}
}

@article{selectiveatt,
  title={Neural mechanisms of selective visual attention},
  author={Desimone, Robert and Duncan, John},
  journal={Annual review of neuroscience},
  volume={18},
  number={1},
  pages={193--222},
  year={1995},
  publisher={Annual Reviews 4139 El Camino Way, PO Box 10139, Palo Alto, CA 94303-0139, USA}
}

@article{fewshot,
  title={Language models are few-shot learners},
  author={Brown, Tom and Mann, Benjamin and Ryder, Nick and Subbiah, Melanie and Kaplan, Jared D and Dhariwal, Prafulla and Neelakantan, Arvind and Shyam, Pranav and Sastry, Girish and Askell, Amanda and others},
  journal={Advances in neural information processing systems},
  volume={33},
  pages={1877--1901},
  year={2020}
}

@book{conceptfirst,
  title={A Study of Thinking},
  author={Jerome S. Bruner, Jacqueline F. Goodnow, George A. Austin},
  publisher={Wiley},
  year={1967}
}

@article{baystwo,
  title={Human-level concept learning through probabilistic program induction},
  author={Lake, Brenden M and Salakhutdinov, Ruslan and Tenenbaum, Joshua B},
  journal={Science},
  volume={350},
  number={6266},
  pages={1332--1338},
  year={2015},
  publisher={American Association for the Advancement of Science}
}

@article{another,
  title={Relational language and the development of relational mapping},
  author={Loewenstein, Jeffrey and Gentner, Dedre},
  journal={Cognitive psychology},
  volume={50},
  number={4},
  pages={315--353},
  year={2005},
  publisher={Elsevier}
}

@article{small,
  title={Object name learning provides on-the-job training for attention},
  author={Smith, Linda B and Jones, Susan S and Landau, Barbara and Gershkoff-Stowe, Lisa and Samuelson, Larissa},
  journal={Psychological science},
  volume={13},
  number={1},
  pages={13--19},
  year={2002},
  publisher={Sage Publications Sage CA: Los Angeles, CA}
}

@article{small2,
  title={Evidence for representations of perceptually similar natural categories by 3-month-old and 4-month-old infants},
  author={Quinn, Paul C and Eimas, Peter D and Rosenkrantz, Stacey L},
  journal={Perception},
  volume={22},
  number={4},
  pages={463--475},
  year={1993},
  publisher={SAGE Publications Sage UK: London, England}
}

@article{contprim,
  title={On priming by a sentence context.},
  author={Stanovich, Keith E and West, Richard F},
  journal={Journal of Experimental Psychology: General},
  volume={112},
  number={1},
  pages={1},
  year={1983},
  publisher={American Psychological Association}
}

@article{reading,
  title={The neural architecture of language: Integrative modeling converges on predictive processing},
  author={Schrimpf, Martin and Blank, Idan Asher and Tuckute, Greta and Kauf, Carina and Hosseini, Eghbal A and Kanwisher, Nancy and Tenenbaum, Joshua B and Fedorenko, Evelina},
  journal={Proceedings of the National Academy of Sciences},
  volume={118},
  number={45},
  pages={e2105646118},
  year={2021},
  publisher={National Acad Sciences}
}

@inproceedings{motion,
  title={Motiongpt: Finetuned llms are general-purpose motion generators},
  author={Zhang, vari and Huang, Di and Liu, Bin and Tang, Shixiang and Lu, Yan and Chen, Lu and Bai, Lei and Chu, Qi and Yu, Nenghai and Ouyang, Wanli},
  booktitle={Proceedings of the AAAI Conference on Artificial Intelligence},
  volume={38},
  number={7},
  pages={7368--7376},
  year={2024}
}

@article{gpt4better,
  title={Human-like intuitive behavior and reasoning biases emerged in large language models but disappeared in ChatGPT},
  author={Hagendorff, Thilo and Fabi, Sarah and Kosinski, Michal},
  journal={Nature Computational Science},
  volume={3},
  number={10},
  pages={833--838},
  year={2023},
  publisher={Nature Publishing Group US New York}
}

@article{collaborative,
  title={Fine-tuning language models to find agreement among humans with diverse preferences},
  author={Bakker, Michiel and Chadwick, Martin and Sheahan, Hannah and Tessler, Michael and Campbell-Gillingham, Lucy and Balaguer, Jan and McAleese, Nat and Glaese, Amelia and Aslanides, John and Botvinick, Matt and others},
  journal={Advances in Neural Information Processing Systems},
  volume={35},
  pages={38176--38189},
  year={2022}
}

@article{firstllmreward1,
  title={Fine-tuning language models from human preferences},
  author={Ziegler, Daniel M and Stiennon, Nisan and Wu, Jeffrey and Brown, Tom B and Radford, Alec and Amodei, Dario and Christiano, Paul and Irving, Geoffrey},
  journal={arXiv preprint arXiv:1909.08593},
  year={2019}
}

@article{firstllmreward2,
  title={Learning to summarize with human feedback},
  author={Stiennon, Nisan and Ouyang, Long and Wu, Jeffrey and Ziegler, Daniel and Lowe, Ryan and Voss, Chelsea and Radford, Alec and Amodei, Dario and Christiano, Paul F},
  journal={Advances in Neural Information Processing Systems},
  volume={33},
  pages={3008--3021},
  year={2020}
}

@article{binz2,
  title={Using cognitive psychology to understand GPT-3},
  author={Binz, Marcel and Schulz, Eric},
  journal={Proceedings of the National Academy of Sciences},
  volume={120},
  number={6},
  pages={e2218523120},
  year={2023},
  publisher={National Acad Sciences}
}

@article{28,
  title={Causal reasoning and large language models: Opening a new frontier for causality},
  author={K{\i}c{\i}man, Emre and Ness, Robert and Sharma, Amit and Tan, Chenhao},
  journal={arXiv preprint arXiv:2305.00050},
  year={2023}
}

@misc{34,
  author       = "McCarthy, J. and Minsky, M. and Rochester, N. and Shannon, C.E.  ",
  title        = "A Proposal for the Dartmouth Summer Research Project on Artificial Intelligence. ",
  year         = "1955",
}

@misc{google,
  author       = "Wei, J. and Zhou, D. ",
  title        = "Language models perform reasoning via Chain of Thought",
  howpublished = "Google AI Blog",
  year         = "2022",
}

@misc{1,
  author       = "New York Times",
  title        = "Pentagon Turns to Silicon Valley for Edge in Artificial Intelligence",
  year         = "2016",
}

@misc{2,
  author       = "PARC",
  title        = "Half-Human, Half-Computer? Meet the Modern Centaur",
  year         = "2023",
}

@misc{3,
  author       = "New York Times",
  title        = "A Case for Cooperation Between Machines and Humans",
  year         = "2016",
}

@misc{4,
  author       = "Garry Kasparov",
  title        = "Garry Kasparov on AI, Chess, and the Future of Creativity",
  year         = "2017",
}

@article{5,
  title={The chess master and the computer},
  author={Kasparov, Garry},
  journal={The New York Review of Books},
  volume={57},
  number={2},
  pages={16--19},
  year={2010}
}

@article{startup,
  title={Using machine learning to demystify startups’ funding, post-money valuation, and success},
  author={Ang, Y. Q. and Chia, A. and Saghafian, S.},
  journal={Springer International Publishing},
  pages={271-296},
  year={2022}
}

@inproceedings{6,
  title={A human-ai collaborative approach for clinical decision making on rehabilitation assessment},
  author={Lee, Min Hun and Siewiorek, Daniel P and Smailagic, Asim and Bernardino, Alexandre and Berm{\'u}dez i Badia, Sergi Berm{\'u}dez},
  booktitle={Proceedings of the 2021 CHI conference on human factors in computing systems},
  pages={1--14},
  year={2021}
}

@article{10,
  title={Ambiguous dynamic treatment regimes: A reinforcement learning approach},
  author={Saghafian, Soroush},
  journal={Management Science},
  volume={70},
  number={9},
  pages={5667--5690},
  year={2024}
}

@article{11,
  title={Comparison of post-transplantation diabetes mellitus incidence and risk factors between kidney and liver transplantation patients},
  author={Munshi, Vidit N and Saghafian, Soroush and Cook, Curtiss B and Werner, K Tuesday and Chakkera, Harini A},
  journal={PloS one},
  volume={15},
  number={1},
  pages={e0226873},
  year={2020},
  publisher={Public Library of Science San Francisco, CA USA}
}

@article{12,
  title={Integrative cell formation and layout design in cellular manufacturing systems},
  author={Saghafian, Soroush and Akbari Jokar, M Reza},
  journal={Journal of Industrial and Systems Engineering},
  volume={3},
  number={2},
  pages={97--115},
  year={2009},
  publisher={Iranian Institute of Industrial Engineering}
}

@book{13,
  title={Dynamic Assignment of Patients to Primary and Secondary Inpatient Units: Is Patience a Virtue?},
  author={Saghafian, Soroush and Kilinc, Derya and  Traub, Stephen},
  year={2024},
  publisher={Cambridge Handbook on Productivity, Efficiency
and Effectiveness in Healthcare, Cambridge University Press}
}

@article{15,
  title={Blink: The power of thinking without thinking},
  author={Gladwell, Malcolm},
  year={2006},
  publisher={Penguin Books London}
}

@article{17,
  title={Ambiguous partially observable Markov decision processes: Structural results and applications},
  author={Saghafian, Soroush},
  journal={Journal of Economic Theory},
  volume={178},
  pages={1--35},
  year={2018},
  publisher={Elsevier}
}

@book{41,
  title={The book of why: the new science of cause and effect},
  author={Pearl, Judea and Mackenzie, Dana},
  year={2018},
  publisher={Basic books}
}

@article{32,
  title={Overcoming algorithm aversion: People will use imperfect algorithms if they can (even slightly) modify them},
  author={Dietvorst, Berkeley J and Simmons, Joseph P and Massey, Cade},
  journal={Management science},
  volume={64},
  number={3},
  pages={1155--1170},
  year={2018},
  publisher={INFORMS}
}

@article{33,
  title={A systematic review of algorithm aversion in augmented decision making},
  author={Burton, Jason W and Stein, Mari-Klara and Jensen, Tina Blegind},
  journal={Journal of behavioral decision making},
  volume={33},
  number={2},
  pages={220--239},
  year={2020},
  publisher={Wiley Online Library}
}

@article{31,
  title={Algorithm aversion: people erroneously avoid algorithms after seeing them err.},
  author={Dietvorst, Berkeley J and Simmons, Joseph P and Massey, Cade},
  journal={Journal of Experimental Psychology: General},
  volume={144},
  number={1},
  pages={114},
  year={2015},
  publisher={American Psychological Association}
}

@article{18,
  title={The newsvendor under demand ambiguity: Combining data with moment and tail information},
  author={Saghafian, Soroush and Tomlin, Brian},
  journal={Operations Research},
  volume={64},
  number={1},
  pages={167--185},
  year={2016},
  publisher={INFORMS}
}

@book{16,
  title={Sources of power: How people make decisions},
  author={Klein, Gary A},
  year={2017},
  publisher={MIT press}
}

@article{14,
  title={Physician in triage versus rotational patient assignment},
  author={Traub, Stephen J and Bartley, Adam C and Smith, Vernon D and Didehban, Roshanak and Lipinski, Christopher A and Saghafian, Soroush},
  journal={The Journal of emergency medicine},
  volume={50},
  number={5},
  pages={784--790},
  year={2016},
  publisher={Elsevier}
}

@article{9,
  title={Heads or tails: The impact of a coin toss on major life decisions and subsequent happiness},
  author={Levitt, Steven D},
  journal={The Review of Economic Studies},
  volume={88},
  number={1},
  pages={378--405},
  year={2021},
  publisher={Oxford University Press}
}

@article{8,
  author       = "Deloitte",
  title        = "How New Human-Machine Collaborations Could Make Government Organizations 
More Efficient.",
journal="Harvard Business Review",
  year         = "2020",
}

@misc{energy,
  author       = "The Register",
  title        = "Energy breakthrough needed to build AGI, says OpenAI boss Altman",
  year         = "2024",
}

@misc{openai,
  author       = "OpenAI",
  title        = "GPT-4",
  howpublished = "https://openai.com/research/gpt-4",
  year         = "2023",
}

@article{intentions,
  title={Training language models to follow instructions with human feedback},
  author={Ouyang, Long and Wu, Jeffrey and Jiang, Xu and Almeida, Diogo and Wainwright, Carroll and Mishkin, Pamela and Zhang, Chong and Agarwal, Sandhini and Slama, Katarina and Ray, Alex and others},
  journal={Advances in neural information processing systems},
  volume={35},
  pages={27730--27744},
  year={2022}
}

@article{values,
  title={Beavertails: Towards improved safety alignment of llm via a human-preference dataset},
  author={Ji, Jiaming and Liu, Mickel and Dai, Josef and Pan, Xuehai and Zhang, Chi and Bian, Ce and Chen, Boyuan and Sun, Ruiyang and Wang, Yizhou and Yang, Yaodong},
  journal={Advances in Neural Information Processing Systems},
  volume={36},
  year={2024}
}

@article{prompt1,
  title={Do prompt-based models really understand the meaning of their prompts?},
  author={Webson, Albert and Pavlick, Ellie},
  journal={arXiv preprint arXiv:2109.01247},
  year={2021}
}

@article{27,
  title={Prospect theory: An analysis of decision under risk},
  author={Kahneman, Daniel and Tversky, Amos},
  booktitle={Econometrica},
volume={47},
  pages={263–292},
  year={1979},
}

@article{36,
  title={Subjective probability: A judgment of representativeness},
  author={Kahneman, Daniel and Tversky, Amos},
  journal={Cognitive psychology},
  volume={3},
  number={3},
  pages={430--454},
  year={1972},
  publisher={Elsevier}
}

@article{26,
  title={Task complexity moderates the influence of descriptions in decisions from experience},
  author={Weiss-Cohen, Leonardo and Konstantinidis, Emmanouil and Speekenbrink, Maarten and Harvey, Nigel},
  journal={Cognition},
  volume={170},
  pages={209--227},
  year={2018},
  publisher={Elsevier}
}

@article{prompt2,
  title={Rethinking the role of demonstrations: What makes in-context learning work?},
  author={Min, Sewon and Lyu, Xinxi and Holtzman, Ari and Artetxe, Mikel and Lewis, Mike and Hajishirzi, Hannaneh and Zettlemoyer, Luke},
  journal={arXiv preprint arXiv:2202.12837},
  year={2022}
}

@book{cogdef,
  title={The Cambridge handbook of computational psychology},
  author={Sun, Ron},
  year={2008},
  publisher={Cambridge University Press}
}

@article{sor,
  title={Algorithm, Human, or the Centaur: How to Enhance Clinical Care?},
  author={Orfanoudaki, Agni and Saghafian, Soroush and Song, Karen and Chakkera, Harini A and Cook, Curtiss},
  year={2022},
  publisher={HKS Working Paper No. RWP22-027}
}

@article{studies,
  title={Out of one, many: Using language models to simulate human samples},
  author={Argyle, Lisa P and Busby, Ethan C and Fulda, Nancy and Gubler, Joshua R and Rytting, Christopher and Wingate, David},
  journal={Political Analysis},
  volume={31},
  number={3},
  pages={337--351},
  year={2023},
  publisher={Cambridge University Press}
}

@article{personality1,
  title={Evaluating and inducing personality in pre-trained language models},
  author={Jiang, Guangyuan and Xu, Manjie and Zhu, Song-Chun and Han, Wenjuan and Zhang, Chi and Zhu, Yixin},
  journal={Advances in Neural Information Processing Systems},
  volume={36},
  year={2024}
}

@article{personality2,
  title={Identifying and manipulating the personality traits of language models},
  author={Caron, Graham and Srivastava, Shashank},
  journal={arXiv preprint arXiv:2212.10276},
  year={2022}
}

@inproceedings{studies2,
  title={Using large language models to simulate multiple humans and replicate human subject studies},
  author={Aher, Gati V and Arriaga, Rosa I and Kalai, Adam Tauman},
  booktitle={International Conference on Machine Learning},
  pages={337--371},
  year={2023},
  organization={PMLR}
}

@article{emotion,
  title={Emotion, cognition, and decision making},
  author={Schwarz, Norbert},
  journal={Cognition \& emotion},
  volume={14},
  number={4},
  pages={433--440},
  year={2000},
  publisher={Taylor \& Francis}
}

@article{bias1,
  title={Large language models show human-like content biases in transmission chain experiments},
  author={Acerbi, Alberto and Stubbersfield, Joseph M},
  journal={Proceedings of the National Academy of Sciences},
  volume={120},
  number={44},
  pages={e2313790120},
  year={2023},
  publisher={National Acad Sciences}
}

@article{conc1,
  title={Augmenting the algorithm: Emerging human-in-the-loop work configurations},
  author={Gr{\o}nsund, Tor and Aanestad, Margunn},
  journal={The Journal of Strategic Information Systems},
  volume={29},
  number={2},
  pages={101614},
  year={2020},
  publisher={Elsevier}
}

@incollection{transfer,
  title={Transfer learning},
  author={Torrey, Lisa and Shavlik, Jude},
  booktitle={Handbook of research on machine learning applications and trends: algorithms, methods, and techniques},
  pages={242--264},
  year={2010},
  publisher={IGI global}
}

@article{conc5,
  title={Comparative Advantage of Humans vs AI in the Long Tail},
  author={Agarwal, Nikhil and Huang, Ray and Moehring, Alex and Rajpurkar, Pranav and Salz, Tobias and Yu, Feiyang},
  journal={American Economic Association},
year={2024}
}

@article{conc4,
  title={Beyond purchase intentions: Mining behavioral intentions of social-network users},
  author={Lihi Idan},
volume       = {40},
  number       = {5},
  pages        = {1111--1132},
  journal={International Journal of Human-Computer Interaction},
year={2022},
doi          = {10.1080/10447318.2022.2132195}
}

@article{conc3,
  title={Will artificial intelligence replace radiologists?},
  author={Langlotz, Curtis P},
  journal={Radiology: Artificial Intelligence},
  volume={1},
  number={3},
  year={2019},
  publisher={Radiological Society of North America}
}

@inproceedings{conc2,
  title={Tackling Face Verification Edge Cases: In-Depth Analysis and Human-Machine Fusion Approach},
  author={Knoche, Martin and Rigoll, Gerhard},
  booktitle={2023 18th International Conference on Machine Vision and Applications (MVA)},
  pages={1--5},
  year={2023},
  organization={IEEE}
}

@article{demons,
  title={Reward learning from human preferences and demonstrations in atari},
  author={Ibarz, Borja and Leike, Jan and Pohlen, Tobias and Irving, Geoffrey and Legg, Shane and Amodei, Dario},
  journal={Advances in neural information processing systems},
  volume={31},
  year={2018}
}

@article{studies3,
  title={Can large language models transform computational social science?},
  author={Ziems, Caleb and Held, William and Shaikh, Omar and Chen, Jiaao and Zhang, Zhehao and Yang, Diyi},
  journal={Computational Linguistics},
  pages={1--55},
  year={2024},
  publisher={MIT Press One Broadway, 12th Floor, Cambridge, Massachusetts 02142, USA~…}
}

@inproceedings{bias2,
  title={Towards understanding and mitigating social biases in language models},
  author={Liang, Paul Pu and Wu, Chiyu and Morency, Louis-Philippe and Salakhutdinov, Ruslan},
  booktitle={International Conference on Machine Learning},
  pages={6565--6576},
  year={2021},
  organization={PMLR}
}

@article{bias3,
  title={Large pre-trained language models contain human-like biases of what is right and wrong to do},
  author={Schramowski, Patrick and Turan, Cigdem and Andersen, Nico and Rothkopf, Constantin A and Kersting, Kristian},
  journal={Nature Machine Intelligence},
  volume={4},
  number={3},
  pages={258--268},
  year={2022},
  publisher={Nature Publishing Group UK London}
}

@book{fairness,
  title={Fairness and machine learning: Limitations and opportunities},
  author={Barocas, Solon and Hardt, Moritz and Narayanan, Arvind},
  year={2023},
  publisher={MIT Press}
}

@article{mancomputer,
  title={Man-computer symbiosis},
  author={Licklider, Joseph CR},
  journal={IRE transactions on human factors in electronics},
  number={1},
  pages={4--11},
  year={1960},
  publisher={IEEE}
}

@inproceedings{ensemble1,
  title={Human-in-the-loop person re-identification},
  author={Wang, Hanxiao and Gong, Shaogang and Zhu, Xiatian and Xiang, Tao},
  booktitle={Computer Vision--ECCV 2016: 14th European Conference, Amsterdam, The Netherlands, October 11--14, 2016, Proceedings, Part IV 14},
  pages={405--422},
  year={2016},
  organization={Springer}
}

@inproceedings{ensemble2,
  title={Human Uncertainty Inference via Deterministic Ensemble Neural Networks},
  author={Cha, Yujin and Lee, Sang Wan},
  booktitle={Proceedings of the AAAI Conference on Artificial Intelligence},
  volume={35},
  number={7},
  pages={5877--5886},
  year={2021}
}

@article{pattern,
  title={Large language models as general pattern machines},
  author={Mirchandani, Suvir and Xia, Fei and Florence, Pete and Ichter, Brian and Driess, Danny and Arenas, Montserrat Gonzalez and Rao, Kanishka and Sadigh, Dorsa and Zeng, Andy},
  journal={arXiv preprint arXiv:2307.04721},
  year={2023}
}

@article{pattern2,
  title={Language is not all you need: Aligning perception with language models},
  author={Huang, Shaohan and Dong, Li and Wang, Wenhui and Hao, Yaru and Singhal, Saksham and Ma, Shuming and Lv, Tengchao and Cui, Lei and Mohammed, Owais Khan and Patra, Barun and others},
  journal={Advances in Neural Information Processing Systems},
  volume={36},
  pages={72096--72109},
  year={2023}
}

@article{klein,
  title={A naturalistic decision making perspective on studying intuitive decision making},
  author={Klein, Gary},
  journal={Journal of applied research in memory and cognition},
  volume={4},
  number={3},
  pages={164--168},
  year={2015},
  publisher={Elsevier}
}

@book{generativebook,
  title={Handbook of pattern recognition and computer vision},
  author={Chen, Chi Hau},
  year={2015},
  publisher={World scientific}
}

@article{kk,
  title={Conditions for intuitive expertise: a failure to disagree.},
  author={Kahneman, Daniel and Klein, Gary},
  journal={American psychologist},
  volume={64},
  number={6},
  pages={515},
  year={2009},
  publisher={American Psychological Association}
}

@article{hogarth,
  title={Educating intuition},
  author={Hogarth, Robin M},
  journal={The University of Chicago},
  year={2001}
}

@article{simone,
  title={What is an “explanation” of behavior?},
  author={Simon, Herbert A},
  journal={Psychological science},
  volume={3},
  number={3},
  pages={150--161},
  year={1992},
  publisher={SAGE Publications Sage CA: Los Angeles, CA}
}

@article{fastslow,
  title={Thinking, fast and slow},
  author={Kahneman, Daniel},
  journal={Farrar, Straus and Giroux},
  year={2011}
}

@article{otte,
  title={Intuition and logic},
  author={Otte, Michael},
  journal={For the learning of mathematics},
  volume={10},
  number={2},
  pages={37--43},
  year={1990},
  publisher={JSTOR}
}

@article{strategic,
  title={The role of intuition in strategic decision making},
  author={Khatri, Naresh and Ng, H Alvin},
  journal={Human relations},
  volume={53},
  number={1},
  pages={57--86},
  year={2000},
  publisher={Sage Publications Sage CA: Thousand Oaks, CA}
}

@article{hbrnew,
  author       = "Laura Huang",
  title        = "When It’s OK to Trust Your Gut on a Big Decision.",
journal="Harvard Business Review",
  year         = "2019",
}

@article{o11,
  title={Evaluation of OpenAI o1: Opportunities and Challenges of AGI},
  author={Zhong, Tianyang and Liu, Zhengliang and Pan, Yi and Zhang, Yutong and Zhou, Yifan and Liang, Shizhe and Wu, Zihao and Lyu, Yanjun and Shu, Peng and Yu, Xiaowei and others},
  journal={arXiv preprint arXiv:2409.18486},
  year={2024}
}

@misc{o12,
  author       = "OpenAI",
  title        = "OpenAI o1 System Card",
howpublished = "https://openai.com/index/openai-o1-system-card/",
  year         = "2024",
}

@article{refprofasked,
  title={Harmonizing Safety and Speed: A Human-Algorithm Approach to Enhance the FDA's Medical Device Clearance Policy},
  author={Zhalechian, Mohammad and Saghafian, Soroush and Robles, Omar},
  journal={Available at SSRN 4863134},
  year={2024}
}

@article{darktriad,
  title={The dark triad of personality: Narcissism, Machiavellianism, and psychopathy},
  author={Paulhus, Delroy L and Williams, Kevin M},
  journal={Journal of research in personality},
  volume={36},
  number={6},
  pages={556--563},
  year={2002},
  publisher={Elsevier}
}

@article{risk1,
  title={Model evaluation for extreme risks},
  author={Shevlane, Toby and Farquhar, Sebastian and Garfinkel, Ben and Phuong, Mary and Whittlestone, Jess and Leung, Jade and Kokotajlo, Daniel and Marchal, Nahema and Anderljung, Markus and Kolt, Noam and others},
  journal={arXiv preprint arXiv:2305.15324},
  year={2023}
}

@article{risk2,
  title={Evaluating frontier models for dangerous capabilities},
  author={Phuong, Mary and Aitchison, Matthew and Catt, Elliot and Cogan, Sarah and Kaskasoli, Alexandre and Krakovna, Victoria and Lindner, David and Rahtz, Matthew and Assael, Yannis and Hodkinson, Sarah and others},
  journal={arXiv preprint arXiv:2403.13793},
  year={2024}
}

@inproceedings{risk3,
  title={A holistic approach to undesired content detection in the real world},
  author={Markov, Todor and Zhang, Chong and Agarwal, Sandhini and Nekoul, Florentine Eloundou and Lee, Theodore and Adler, Steven and Jiang, Angela and Weng, Lilian},
  booktitle={Proceedings of the AAAI Conference on Artificial Intelligence},
  volume={37},
  number={12},
  pages={15009--15018},
  year={2023}
}

@inproceedings{sor1,
  title={The Analytics Science Behind {ChatGPT}: Human, Algorithm, or a Human-Algorithm Centaur?},
  author={Saghafian, Soroush},
  booktitle={Public Impact Analytics Science (PIAS) Blog},
  volume={},
  number={},
  pages={},
  year={2023}
}

@article{sor2,
  title={Patient streaming as a mechanism for improving responsiveness in emergency departments},
  author={Saghafian, Soroush and Hopp, Wallace J and Van Oyen, Mark P and Desmond, Jeffrey S and Kronick, Steven L},
  journal={Operations Research},
  volume={60},
  number={5},
  pages={1080--1097},
  year={2012},
  publisher={INFORMS}
}

@article{reasoning,
  title={Teaching algorithmic reasoning via in-context learning},
  author={Zhou, Hattie and Nova, Azade and Larochelle, Hugo and Courville, Aaron and Neyshabur, Behnam and Sedghi, Hanie},
  journal={arXiv preprint arXiv:2211.09066},
  year={2022}
}

@article{wei2022chain,
  title={Chain-of-thought prompting elicits reasoning in large language models},
  author={Wei, Jason and Wang, Xuezhi and Schuurmans, Dale and Bosma, Maarten and Xia, Fei and Chi, Ed and Le, Quoc V and Zhou, Denny and others},
  journal={Advances in neural information processing systems},
  volume={35},
  pages={24824--24837},
  year={2022}
}

@article{planning,
  title={LLMs Still Can't Plan; Can LRMs? A Preliminary Evaluation of OpenAI's o1 on PlanBench},
  author={Valmeekam, Karthik and Stechly, Kaya and Kambhampati, Subbarao},
  journal={arXiv preprint arXiv:2409.13373},
  year={2024}
}
\end{document}